\documentclass[conference]{IEEEtran}
\IEEEoverridecommandlockouts
\usepackage{cite}
\usepackage{textcomp}
\usepackage{xcolor}
\def\BibTeX{{\rm B\kern-.05em{\sc i\kern-.025em b}\kern-.08em
    T\kern-.1667em\lower.7ex\hbox{E}\kern-.125emX}}

\usepackage{float}
\usepackage[inline]{enumitem}
\usepackage{array}
\usepackage{tabularx}
\usepackage{booktabs}
\usepackage{makecell}
\usepackage{pifont}
\newcommand{\cmark}{\ding{51}}
\newcommand{\xmark}{\ding{55}}

\usepackage{amsmath,amssymb,amsfonts}

\usepackage[endLComment=]{algpseudocodex}
\usepackage[ruled]{algorithm}

\usepackage{comment}
\usepackage{url}

\usepackage{xspace}
\usepackage{graphicx}
\usepackage{colortbl}
\usepackage{listings}
\usepackage{placeins}
\usepackage{shuffle}
\usepackage{subcaption}
\usepackage{ifthen}
\usepackage{xurl}
\usepackage{hyperref}

\setlist[itemize]{nosep, leftmargin=1em}
\setlist[description]{nosep}
\setlist[enumerate]{nosep}
\newcommand{\nomad}{\textsc{Nomad}\xspace}

\begin{document}
\title{NOMAD: Generating Embeddings for \\ Massive Distributed Graphs}
\author{
    \IEEEauthorblockN{Aishwarya Sarkar\IEEEauthorrefmark{1}, Sayan Ghosh\IEEEauthorrefmark{2}, Nathan R. Tallent\IEEEauthorrefmark{2}, Ali Jannesari\IEEEauthorrefmark{1}}
    \IEEEauthorblockA{\IEEEauthorrefmark{1}Iowa State University, Ames, IA, USA, Email: \{asarkar1, jannesari\}@iastate.edu}
    \IEEEauthorblockA{\IEEEauthorrefmark{2}Pacific Northwest National Laboratory, Richland, WA, USA, Email: \{sayan.ghosh, tallent\}@pnnl.gov}
}
% \and
% \and
% \and
% \and
% \and

\maketitle

\begin{abstract}
Successful machine learning on graphs or networks requires embeddings that not only represent nodes and edges as low-dimensional vectors but also preserve the graph structure. Established methods for generating embeddings require flexible exploration of the entire graph through repeated use of random walks that capture graph structure with samples of nodes and edges. These methods create scalability challenges for massive graphs with millions-to-billions of edges because single-node solutions have inadequate memory and processing capabilities. 

We present \nomad, a distributed-memory graph embedding framework using the Message Passing Interface (MPI) for distributed graphs. \nomad implements proximity-based models proposed in the widely popular LINE (Large-scale Information Network Embedding) algorithm. We propose several practical trade-offs to improve the scalability and communication overheads confronted by irregular and distributed graph embedding methods, catering to massive-scale graphs arising in web and science domains. \nomad demonstrates median speedups of 10\slash 100$\times$ on CPU-based NERSC Perlmutter cluster relative to the popular reference implementations of multi-threaded LINE and node2vec, 35--76$\times$ over distributed PBG, and competitive embedding quality relative to LINE, node2vec, and GraphVite, while yielding 12--370$\times$ end-to-end speedups on real-world graphs.

% and baseline evaluations yield 12--370$\times$ improvements in end-to-end training times for complex real-world graphs. 
% \note{what is the best speedup or you can mention a range}

%Node embedding generation is a fundamental requirement of modern graph-based learning, powering tasks such as node classification, link prediction, recommendation, and large-scale knowledge graph reasoning. However, for web-scale and scientific graphs, such as those in biology and chemistry, containing millions to billions of nodes and edges, embedding generation becomes a major systems bottleneck due to irregular memory access patterns, high communication overhead, and longer end-to-end training time.
\end{abstract}

\section{Introduction}\label{sec:intro}
\begin{table*}[t]
\centering
\scriptsize
\caption{\small Comparison of NOMAD with existing graph sampling and embedding systems.}
\setlength{\tabcolsep}{4pt}
\renewcommand{\arraystretch}{1.15}
\resizebox{\textwidth}{!}{%
\begin{tabular}{|l|c|c|c|c|c|c|c|c|l|}
\hline
\textbf{Framework} & \textbf{CPU} & \textbf{GPU} & \textbf{Distributed} & \textbf{Sampling} & \textbf{Training} & \textbf{Scale Up} & \textbf{Scale Out} & \textbf{Partitioned graph} & \textbf{Notes} \\
\hline
\hline
NOMAD (this work) & \cmark & \xmark & \cmark & \cmark & \cmark & \cmark & \cmark & \cmark & MPI collective communication and synchronization \\
\hline
\hline
PyTorch-BigGraph~\cite{lerer2019pytorch} & \cmark & \xmark & \cmark & \cmark & \cmark & \cmark & \cmark & \cmark & Parameter-server architecture with shared parameter synchronization \\
\hline
Fast-Node2vec~\cite{zhou2018efficient} & \cmark & \xmark & \cmark & \cmark & \cmark & \cmark & \cmark & \cmark & Pregel-like iterative BSP \\
\hline
GraphVite~\cite{zhu2019graphvite} & \cmark & \cmark & \xmark & \cmark & \cmark & \cmark & \xmark & \xmark & Hybrid, OpenMP threads for sampling; GPU-based training \\
\hline
Grape~\cite{cappelletti2023grape} & \cmark & \xmark & \xmark & \cmark & \cmark & \cmark & \xmark & \xmark & Thread-level SIMD parallelism \\
\hline
CuGraph~\cite{cugraph} & \xmark & \cmark & \cmark & \cmark & \xmark & \cmark & \cmark & \cmark & Dask-based multi-GPU random walk sampling (no training) \\
\hline
NextDoor~\cite{jangda2021accelerating} & \xmark & \cmark & \xmark & \cmark & \xmark & \cmark & \xmark & \xmark & Multi-GPU vertex-based sampling (no training) \\
\hline
C-SAW~\cite{pandey2020c} & \xmark & \cmark & \xmark & \cmark & \xmark & \cmark & \xmark & \cmark & Multi-GPU sampling (partitions graph across GPUs, no training) \\
\hline
gSampler~\cite{gong2023gsampler} & \xmark & \cmark & \xmark & \cmark & \xmark & \cmark & \xmark & \xmark & Multi-GPU sampling (no training) \\
\hline
\end{tabular}%
}
\label{tab:system-comparison}
\end{table*}
Graphs are ubiquitous in modeling complex systems and representing entity interactions to uncover a domain's structural information. 
Graph representation learning broadly spans shallow transductive methods (proximity-based) that learn directly from graph structure and inductive message-passing models (graph neural networks (GNNs)) that additionally rely on node features and supervision (labels)~\cite{hamilton2017representation,chen2019exploiting}. Proximity-based methods such as LINE~\cite{tang2015line}, DeepWalk~\cite{perozzi2014deepwalk}, and node2vec~\cite{grover2016node2vec} belong to the former class, learning node representations from sampled pairs derived from edge lists or breadth\slash depth-first traversals or that generate random sequences from a probability distribution. Walk-based methods such as DeepWalk and node2vec derive training pairs from multi-hop co-occurrence context, whereas LINE directly optimizes sampled proximity pairs (faster on massive graphs). 
%Walk-based methods such as DeepWalk and node2vec derive training pairs from multi-hop co-occurrence context, whereas LINE directly optimizes sampled proximity pairs; these objectives are closely related under a unifying matrix-factorization view~\cite{qiu2018network}. 
For many real-world large scientific and web graphs, attributes and labels may be incomplete, noisy, heterogeneous, or unavailable making GNN training less effective or harder to apply directly~\cite{kiss2024unsupervised,wang2022heterogeneous,na2025heterogeneous,taguchi2021graph,wu2015protein}. In such cases, structure-only methods are a more practical choice for unsupervised learning and can also provide useful initial node representations for downstream supervised GNNs, when informative node features are missing\slash unreliable.
% Although traditional graph analytics learns (without supervision) from explicit graph representations and unlabeled data, modern graph-based deep learning methods such as Graph Neural Networks (GNNs) rely on the iteratively-refined vector representations (i.e., embeddings) that encode  graph structure through nodes/edges, node/edge properties, and difficult-to-obtain ground-truth labels.
%Although traditional graph analytics aims to solve unsupervised problems to learn from unlabeled data at scale, while modern graph-based deep learning methods such as Graph Neural Network (GNN) relies on the initial vector representations (i.e., embeddings) of nodes\slash edges encoding the graph structure\slash features and difficult-to-obtain ground-truth labels for improving the accuracy of learning the node representations.

% When generating embeddings GNNs can uncover global structural context by incorporating multi-hop neighborhood information.
% However in the typical unsupervised setting with noisy data, traditional shallow few-hop methods provide the necessary initialization baseline.
% Hence, few-hop-based node embedding is a practical prerequisite of modern graph-based learning, empowering tasks such as node classification, link prediction, recommendation, and large-scale knowledge graph reasoning. 
Traditional dimensionality-reduction approaches are typically quadratic in the number of vertices and therefore intractable for medium-to-large graphs~\cite{tenenbaum2000global, belkin2003laplacian, cox2008multidimensional, pezzotti2016hierarchical}. For large graphs with millions to billions of nodes and edges, such as the web-scale and scientific graphs found in biology and chemistry, embedding generation becomes a major systems bottleneck due to irregular memory access, limited locality, and high communication overhead. This cost stems from repeated random-walk exploration and training on the resulting sampled pairs, which expose irregular communication and load imbalance at scale. Random walks remain a standard mechanism for extracting structural information for graph embedding, irrespective of specific walk biases used to generate the node sequences~\cite{vital2024comparing}. However, although state-of-the-art shallow embedding systems exploit multithreading and other within-node optimizations, they remain constrained by node memory capacity, and methods that preserve global network properties often require synchronization patterns that do not scale well across distributed memory~\cite{goyal2018graph,zhu2019graphvite}.

Prior efforts on distributed graph embeddings either focus on accelerating a part of the pipeline (like random walks~\cite{willich2025scalerunner}) or rely on parameter-server architecture (to manage and synchronize shared model parameters)~\cite{lerer2019pytorch} with limited scalability potential across nodes. Parallelism based on graph partitions alone cannot scale this problem without specific interventions since inherent irregularities in the graph partitions  cause load imbalance (partitioning is NP-hard~\cite{bui1992finding}). At scale, graph-induced access patterns require both memory capacity and sufficient compute concurrency, making single-node scale-up approaches insufficient, necessitating distributed scale-out over partitioned graphs. Capacity alone is not enough: without proportional bandwidth and compute, disaggregated or pooled memory cannot effectively sustain embedding generation for these irregular workloads~\cite{zhang2023rethinking}.

Table~\ref{tab:system-comparison} summarizes the design space of representative graph embedding systems~\cite{lerer2019pytorch, zhou2018efficient, zhu2019graphvite, cappelletti2023grape, cugraph, jangda2021accelerating, pandey2020c, gong2023gsampler} across relevant dimensions, including hardware platform, end-to-end sampling and training support, and scale-up\slash scale-out capabilities. 
Single-node CPU/GPU based embedding generation can drive node throughput through optimized memory layouts, concurrency, host-device overlap, and cache-efficient sampling~\cite{cappelletti2023grape, zhu2019graphvite}. But such optimizations fall apart on distributed-memory considering partitioned graphs, which must reconcile with irregular data intensive graph exploration (either through \emph{structure-preserving} random walks or weighted edge sampling), coordination with training and managing synchronization overheads. 
%Single-node CPU systems~\cite{cappelletti2023grape} improve within-node throughput through optimized memory layouts, thread-level parallelism, and cache-efficient sampling, but remain constrained by fixed node memory capacity and do not address partitioned execution across nodes. GPU-based and hybrid CPU--GPU systems~\cite{zhu2019graphvite} can further accelerate local training and sample generation on a single node, but extending them to distributed memory is challenging as random-walk sampling is irregular, embedding updates require frequent access to partitioned state, and coordinating sampling and training across CPUs and GPUs introduces additional data movement, synchronization, and complexity.  
Consequently, existing distributed GPU approaches consider either sampling or training, but not both as part of the same system.
We propose \nomad, a distributed-memory scalable
proximity-based end-to-end graph embedding generator (that learns from sampled positive/negative pairs similar to LINE~\cite{tang2015line}, DeepWalk~\cite{perozzi2014deepwalk}, and node2vec~\cite{grover2016node2vec}). \nomad proposes several communication-avoiding and synchronization trade-offs while preserving acceptable embedding accuracy.
%\nomad instead targets end-to-end proximity-based embedding learned from sampled positive/negative pairs (similar to LINE, DeepWalk, and node2vec) on partitioned graphs, where graph data, sampled pairs, and embedding parameters are distributed across processing elements, making communication, synchronization, and cross-partition irregularity first-order concerns.
%NOMAD mitigates the startup and memory issues affecting serial\slash multi-threaded graph embedding generation.
%To avoid the complexity of reconciling hybrid programming models (e.g., process and thread based), \nomad adopts a process-based approach. It makes no assumption on the number of partitions or the node configurations.
For wide portability, \nomad uses MPI collectives to efficiently handle non-uniform communication volumes across non-uniform process sets. 
Our contributions are:
\begin{enumerate}[left=0pt,itemsep=0pt]
    \item Distributed-memory implementation of the popular second-order proximity model of LINE. 
    \item Several performance\slash quality trade-offs for distributed random walk exploration that enable \nomad to enhance communication\slash synchronization efficiency.
    \item Detailed quality and performance evaluation on NERSC Perlmutter across diverse large-scale graphs, showing up to $370\times$ speedup over strong single-node baselines and $35$--$76\times$ speedup over distributed-memory baselines, and competitive embedding quality. 
    %across processes and proposed variants, while matching or improving embedding quality.
    % \item Detailed quality and performance assessments on NERSC Perlmutter using diverse large-scale input graphs, yielding up to 100$\times$ improvement in execution times relative to the state-of-the-art and 12--370$\times$ improvements across processes and proposed variants.
\end{enumerate}
The paper is organized as follows: \S\ref{sec:background} reviews proximity-based embedding and distributed tradeoffs, \S\ref{sec:related} discusses related work, \S\ref{sec:methods} presents \nomad's distributed methodology, and \S\ref{sec:results} reports quality and performance evaluations.
% The paper is organized as follows: \S\ref{sec:background} provides background on proximity-based embedding models\slash scalability challenges, \S\ref{sec:overview} outlines \nomad's high-level system design\slash tradeoffs, \S\ref{sec:methods} presents the distributed variants of \nomad, and \S\ref{sec:results} presents detailed performance\slash quality analysis.

% methods-prelim
\section{Background}\label{sec:background}
\begin{figure}[!ht]
  \centering
\includegraphics[width=\linewidth]{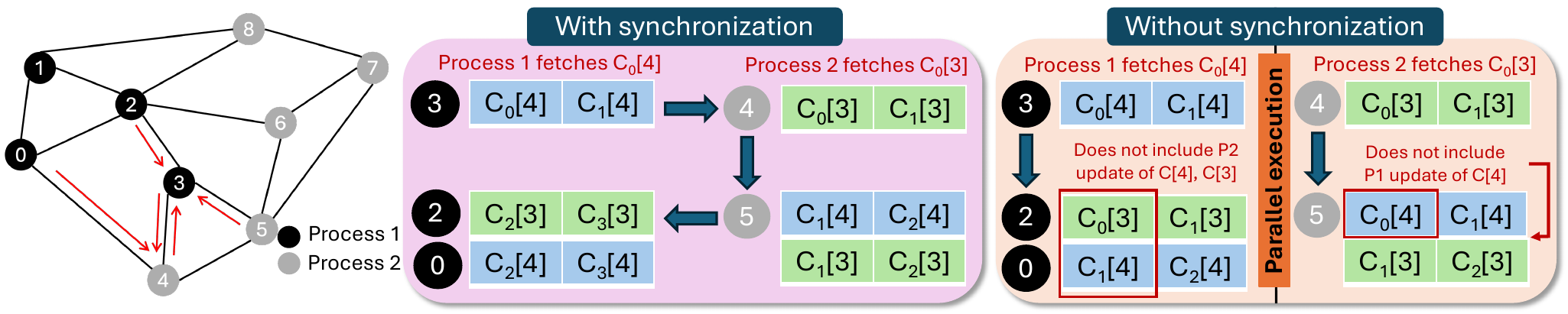}
  \caption{\small With sync: processes observe up-to-date context. Without sync: delayed updates result in stale context, affecting quality.}
%  \caption{With synchronization, processes observe up-to-date context vectors, ensuring correct propagation of updates. Without synchronization, delayed updates result in stale context vectors being reused across processes, illustrating dependency violations that degrade embedding quality.}
  \label{fig:stale}
\end{figure}

Proximity-based embedding models learn low-dimensional vertex representations by optimizing over large collections of independently sampled training pairs of vertices of an arbitrary graph $G = (V, E)$, treating these pairs as the fundamental unit of parallelism. A \emph{positive pair} $(u,v): G = (V, E) \; s.t.\; \{u, v\} \in E \mid u \in V  \;and\; v \in V$ denotes two vertices that are structurally similar (or, closer), typically extracted from local graph neighborhoods or co-occurrence within short random walks, while a \emph{negative pair} $(u,v^-): G = (V, E) \; s.t.\; \{u, v^{-}\} \notin E \mid u \in V  \;and\; v^{-} \in V$ puts the target vertex $u$ against a randomly sampled vertex $v^-$ that does not appear in the same context, providing a stabilizing signal during training~\cite{mikolov2013distributed}. Each node is associated with two low-dimensional vectors, a \emph{vertex} embedding $U$ and a \emph{context} embedding $C$, whose dot product determines similarity.
% Every node consists of two low-dimensional vectors, called \emph{vertex} and \emph{context} that encodes the embedding (denoted as $U$ and $C$), and projection of vertex into context (via dot product) of two arbitrary nodes\slash vertices determines similarity. 
At scale, embedding computation itself is lightweight, consisting primarily of vector dot products and updates. DeepWalk and node2vec treat random walks as ``sentences'' and optimize skip-gram-style objectives over the resulting co-occurrences, while LINE trains on edge- or proximity-derived pairs with similar updates~\cite{mikolov2013efficient,tang2015line}. The dominant cost is instead generating and exchanging sufficient positive pairs under irregular traversal, skewed degree distributions, and limited locality. In distributed memory, many pairs span partitions and require remote embedding access (naive synchronization is not scalable), so practical systems must exploit pair-level parallelism while tolerating bounded staleness and amortizing communication across batches. 
We consider a graph $G=(V,E)$ distributed across $p$ processes. Each process $p$ owns a disjoint subset of vertices $V_p \subset V$ and their associated edges $E_p$. Our default distribution is 1D vertex-based, so each process owns roughly $|V|/p$ vertices and their edges. To improve file I/O, we convert the input graph from native ASCII to binary CSR and apply a first-fit heuristic to seek edge-balanced 1D partitions, falling back to the standard distribution if balancing fails within one iteration. \nomad enforeces an owner-computes model: each embedding is updated only by its owner, while remote processes exchange embeddings through point-to-point messages that are later incorporated into local training. Because training data is sampled independently from each process's partition, updates proceed asynchronously with periodic communication and synchronization. Propagating every context or vertex update immediately to all remote copies would incur prohibitive overhead and introduce circular dependencies at scale. \nomad therefore permits bounded staleness (Fig.~\ref{fig:stale}), allowing processes to continue training with fetched remote context vectors and reconcile updates periodically.

\section{Related Work}\label{sec:related} 
PecanPy~\cite{liu2021pecanpy} and GRAPE~\cite{cappelletti2023grape} improve within-node sampling and training efficiency, while GraphVite~\cite{zhu2019graphvite} combines CPU sampling with GPU training for high single-node throughput. Unlike \nomad, these systems address scale-up and not scale-out, and benefit from co-located memory and avoid the communication, synchronization, and cross-partition irregularity that become first-order bottlenecks for large graphs where the graph and training must be distributed across nodes. We therefore discuss these systems as related work rather than primary baselines for distributed-memory scale-out.
PyTorch-BigGraph (PBG)~\cite{lerer2019pytorch} is the closest distributed baseline to \nomad in scale-out capability and is therefore included in our evaluation. It uses a parameter-server architecture with asynchronous updates over edge buckets, avoiding random-walk sampling and shifting the dominant cost to parameter synchronization rather than sampled-pair generation~\cite{recht2011hogwild,renz2022nups}.
% PyTorch-BigGraph (PBG)~\cite{lerer2019pytorch} is the closest distributed baseline to \nomad in terms of scale-out capability (Table ~\ref{tab:system-comparison}). PBG partitions embeddings and schedules work over edge buckets using a parameter-server architecture with asynchronous updates. Unlike \nomad, it avoids random walk sampling altogether and trains directly on the edge list, shifting the dominant cost from sampled pair generation to parameter synchronization~\cite{recht2011hogwild,renz2022nups}.
% , as it scales embedding training across compute nodes by partitioning embeddings and scheduling work over edge buckets, using a parameter-server architecture with asynchronous updates. Unlike walk-based methods, PBG avoids random walk sampling altogether and trains directly on the edge-list, shifting the scalability challenge from sampling to parameter synchronization.
% Communication is reduced through partition locking and reuse, but global parameters are still synchronized lazily, reintroducing update staleness~\cite{recht2011hogwild} and sensitivity to skewed access patterns~\cite{renz2022nups}. 
% PBG primarily optimizes distributed model updates, whereas \nomad targets the end-to-end cost of distributed walk generation and sampling coordination, which induces severe synchronization overhead, causing collective communication to be predominant  (Table~\ref{comm-analysis}). 
Fast-Node2Vec~\cite{zhou2018efficient} targets the node2vec objective using biased second-order random walks within a Pregel-like bulk-synchronous execution model~\cite{malewicz2010pregel}. As a result, its scalability is strongly shaped by the cost of maintaining biased walk state and by communication associated with high-degree vertices, which differs from the first-order weighted walks used by \nomad solely to construct positive pairs.  Therefore, we do not treat it as a direct baseline for the distributed-memory LINE-style regime targeted in this paper.
% Fast-Node2Vec~\cite{zhou2018efficient} is another distributed-memory implementation of node2vec built on a Pregel-like~\cite{malewicz2010pregel} graph computation framework (using a server-client architecture, where the server coordinates the work across clients in a bulk-synchronous manner) that accelerates node2vec by eliminating precomputed alias tables and performing on-the-fly transition sampling with improved cache locality, significantly reducing memory overhead and preprocessing cost compared to Spark-based implementations.
Other distributed graph representation systems target different objectives from \nomad. For example, distributed linear-algebra-based approaches~\cite{ranawaka2024scalable} reformulate embedding generation with sparse matrix kernels rather than stochastic sampled-pair generation, while information-oriented walk systems such as DistGER~\cite{fang2023distributed} use entropy or correlation-guided walks instead of fixed-length proximity-preserving walks. These systems address different algorithmic and systems bottlenecks from those studied in \nomad, and are not directly comparable to systems that preserve LINE\slash node2vec-style proximity semantics, which is adopted in \nomad's owner-computes paradigm (see \S\ref{sec:background}).
% Other distributed graph representation approaches target different embedding objectives altogether. For example, distributed linear-algebra-based methods~\cite{ranawaka2024scalable} reformulate embedding generation with sparse matrix kernels (SDDMM and SpMM) rather than stochastic walk generation, sampling variability, or collective synchronization behavior central to proximity-based methods studied in \nomad. Likewise, information-oriented random-walk systems such as DistGER~\cite{fang2023distributed} guide walks using entropy or correlation convergence rather than proximity-based methods such as DeepWalk or node2vec, which rely on fixed-length random walks to preserve proximity semantics. Information-oriented walks additionally require maintaining path-dependent state across steps, fundamentally altering the walk distribution and the resulting sample pool. These systems address different algorithmic and systems bottlenecks from those studied in \nomad, and are not directly comparable to systems that preserve LINE\slash node2vec-style proximity semantics, which is adopted in \nomad's owner-computes paradigm (see \S\ref{sec:overview}).
% \subsection{Sampling-only systems}

% \noindent \textbf{Sampling-only systems}
A separate line of work accelerates graph sampling without embedding training. CuGraph~\cite{cugraph}, NextDoor~\cite{jangda2021accelerating}, C-SAW~\cite{pandey2020c}, and gSampler~\cite{gong2023gsampler} focus on random walk or neighborhood sampling, but do not provide an end-to-end embedding system. 
% A separate line of work accelerates graph sampling without performing embedding training. CuGraph~\cite{cugraph}, NextDoor~\cite{jangda2021accelerating}, C-SAW~\cite{pandey2020c}, and gSampler~\cite{gong2023gsampler} focus on single-node or multi-GPU random walk or neighborhood sampling, but do not jointly optimize sampling and embedding training. 
For large graphs, sampling alone is insufficient as an end-to-end baseline, as decoupling sampling from training introduces intermediate copies, redistributions, and coarse-grain synchronization. \nomad instead co-designs sampling and training for distributed-memory embedding generation.

\section{Methods}\label{sec:methods}
%\subsection{Problem Definition}
%\subsection{Distributed Graph Partitioning Model}
In this section, we discuss the design of distributed training (\S\ref{method-training}), generating random samples for training \nomad (\S\ref{method-random-sampling}) and performance\slash quality considerations leading to several variants of distributed random walks (\S\ref{method-remote-variants}).

\subsection{Distributed Training}\label{method-training}
Training proceeds in batches (Algorithm~\ref{alg:nomad_simplified}). For each batch, process $p$ samples positive node pairs $\mathcal{P}$ such that $(u,v)\in\mathcal{P}$ satisfies $\mathrm{owner}(u)=p$ (Line~\ref{alg:line:pos-sample}); thus $u$ is always local, while $v$ may be local or remote. For each positive pair, \nomad draws $K$ negative samples from the global degree-based distribution ($d(v)$ denote the out degree of node $v$):
\begin{align}
    q(v) \;=\; \frac{d(v)^{\alpha}}{\sum_{x \in V} d(x)^{\alpha}}, \qquad \alpha = 0.75
\end{align}
following standard word-embedding practice~\cite{mikolov2013distributed}. A negative pair is formed as $(u,v^-)$ with $v^- \sim q(\cdot)$ (Line~\ref{alg:line:neg-sample}), biasing negatives toward high-degree nodes to improve stability and convergence on skewed graphs. Negative pairs are sampled independently of partition boundaries and may therefore involve remote vertices. By requiring only the target node $u$ to be locally owned, \nomad maximizes asynchrony across processes and avoids conflicting target-embedding updates; sampled remote vertices are buffered locally so communication can be batched and amortized across batches.
% Training proceeds in batches (see Algorithm~\ref{alg:nomad_simplified}). For each batch, process $p$ samples a batch of positive node pairs $\mathcal{P}$, such that $(u,v) \in \mathcal{P}$ satisfies $\mathrm{owner}(u)=p$ (Line~\ref{alg:line:pos-sample}). The target node $u$ is therefore always locally owned, while the context node $v$ may be either local or remote. For each positive pair, \nomad additionally draws $K$ negative samples from a global degree-based distribution. If $d(v)$ denote the out degree of node $v$, negative nodes are sampled from the distribution: 
% \begin{align}
%     q(v) \;=\; \frac{d(v)^{\alpha}}{\sum_{x \in V} d(x)^{\alpha}}, \qquad \alpha = 0.75
% \end{align}
% using the standard distribution introduced in word embedding models~\cite{mikolov2013distributed}. For a given source node $u$, a negative pair is then formed as $(u, v^-)$ where $v^- \sim q(\cdot)$ (Line ~\ref{alg:line:neg-sample}). This strategy biases negative samples toward high-degree nodes, improving training stability and convergence in large, skewed graphs. In \nomad, negative pairs are selected independent of partition boundaries and may correspond to remote vertices. By constraining each sampled pair to have only the target (source, $u$) node locally owned, we maximize asynchrony across processes, avoiding conflicting updates to the target embeddings. Sampled remote vertices are buffered locally during pair generation, enabling subsequent communication to be batched and amortized across batches.
\begin{algorithm}[t]
{\scriptsize
\caption{\footnotesize \textsc{\textbf{NOMAD: Distributed Node Embedding}} \\
\textbf{Inputs:} $G_p=(V_p,E_p)$ portion of the (undirected) graph $G$ on Process $p$, containing ``local'' and ``ghost'' vertices (not owned by process $p$) that have an edge with a vertex $v \in V_p$; batch size $B$; \#negatives per positive pair: K; $\lambda$: weight decay; $\eta$: learning rate; $\alpha_{\text{neg}}$: negative weight;  \\
\textbf{Output:} Embeddings $\mathbf{U}$ and $\mathbf{C}$
}
\label{alg:nomad_simplified}
\begin{algorithmic}[1]
    \State Initialize $\mathbf{U}$ and $\mathbf{C}$
    \For{$b$ in \textit{range(\texttt{\#num\_batches)}}}  \label{alg:line:before-training-loop}
        \State $\mathcal{P}\gets \Call{GeneratePairs}{G_p,B}$\Comment{returns $(u,v)$ with $\mathrm{owner}(u)=p$}\label{alg:line:pos-sample}
        \State $\mathcal{Q}\gets \emptyset$ \Comment{local buffer to store remote $v$ sampled}
        \LComment{local updates}
        \ForAll{$(u,v)\in \mathcal{P}$}\label{alg:line:local-update-begin}
            \If{$\mathrm{owner}(v)=p$}
                % \State $s \gets \sigma(U[u], C[v])$
                \State $g \gets \eta\,(1 - \sigma(U[u], C[v]))$
                \State $\Call{LineUpdate}{U[u], C[v], g}$ \Comment{local positive updates}\label{alg:pos-updates-local}
            \Else
                \State $\mathcal{Q}\gets \mathcal{Q}\cup\{v\}$ \Comment{store remote v}\label{alg:defer-pos-remote}
            \EndIf
            \For{$j=1$ to $K$} \Comment{negative sampling (per positive)}\label{alg:line:neg-sample}
                \State Sample $v^- \text{ from } q(\cdot)$ \Comment{$q(v) \propto d(v)^{0.75}$}\label{alg:line:neg-sample}
                \If{$\mathrm{owner}(v^-)=p$}
                    % \State $s \gets \sigma(U[u], C[v^-])$
                    \State $g \gets -\eta\,\alpha_{\text{neg}}\sigma(U[u], C[v^-])$
                    \State $\Call{LineUpdate}{U[u], C[v^-]}$ \Comment{local negative updates}\label{alg:neg-updates-local}
                \Else
                    \State $\mathcal{Q}\gets \mathcal{Q}\cup\{v^-\}$ \Comment{store remote $v^-$}\label{alg:defer-neg-remote}
                \EndIf
            \EndFor
        \EndFor\label{alg:line:local-update-end}
        \LComment{remote updates}
        \State $\mathbf{Z}\gets \Call{MPI\_Alltoallv}{\mathcal{Q}}$ \Comment{fetch remote context vectors $\mathbf{C}[v]$}\label{alg:mpi-a2a-fetch}
        \State $\Delta \gets \emptyset$\label{alg:accumulate-delta-begin}
        \ForAll{$v \in \mathcal{Q}$} \Comment{remote positive updates}
            \State $g \gets \eta\,(1 - \sigma(U[u], Z[v]))$
            \State $\delta \gets \Call{LineUpdate}{U[u], Z[v], g}$ \Comment{update U}
            \State $\Delta[v] \gets \Delta[v] + \delta$ \Comment{compute $\delta$ for remote C}
        \EndFor
        \ForAll{$v^- \in \mathcal{Q}$} \Comment{remote negative updates}
            \State $g \gets -\eta\,\alpha_{\text{neg}}\sigma(U[u], Z[v^-])$
            \State $\delta \gets \Call{LineUpdate}{U[u], Z[v^-], g}$ \Comment{update U}
            \State $\Delta[v^-] \gets \Delta[v^-] + \delta$ \Comment{compute $\delta$ for remote $C$}
        \EndFor\label{alg:accumulate-delta-end}
        \State $\Call{MPI\_Alltoallv}{\Delta}$\Comment{send accumulated $\Delta$ for updating remote $C$}\label{alg:line:mpi-a2a-send-delta}
    \EndFor
    \State
    \Procedure{LineUpdate}{$U, C, g$}\label{alg:line:emb-update-begin}
    \State $\mathbf{U}_0 \gets U$
    \State $U \gets U + g \cdot C - \eta \cdot \lambda \cdot U$
    \If{local update:}
    \State $C \gets C + g \cdot \mathbf{U}_0 - \eta \cdot \lambda \cdot C$
    \Else 
    \State $\delta \gets g \cdot \mathbf{U}_0 - \eta \cdot \lambda \cdot Z$ \Comment{delta for remote $C$}
    \State \Return $\delta$     
    \EndIf
    \EndProcedure\label{alg:line:emb-update-end}
\end{algorithmic}
}
\end{algorithm}
%\subsection{Local Embedding Updates}
For training pairs whose target and context vertices are owned by the same process, updates are applied immediately without communication or synchronization (Line~\ref{alg:line:local-update-begin}--\ref{alg:line:local-update-end}). For a sampled pair $(u,v)$ with $\mathrm{owner}(u)=\mathrm{owner}(v)=p$, process $p$ computes the gradients with respect to $\mathbf{U}[u]$ and $\mathbf{C}[v]$ and updates them locally; locally owned negative samples are handled identically. These updates follow stochastic gradient descent (Line~\ref{alg:line:emb-update-begin}--\ref{alg:line:emb-update-end}):
% For training pairs whose target and context vertices are owned by the same process, embedding updates are performed immediately (without communication\slash synchronization) (Line~\ref{alg:line:local-update-begin}--\ref{alg:line:local-update-end}). Specifically, for a sampled positive pair $(u,v)$ such that $\mathrm{owner}(u)=\mathrm{owner}(v)=p$, process $p$ computes the gradient of the loss with respect to $\mathbf{U}[u]$ and $\mathbf{C}[v]$ and applies the updates immediately. Locally owned negative samples are handled in the same manner. Local updates follow a stochastic gradient descent update (Line~\ref{alg:line:emb-update-begin}--\ref{alg:line:emb-update-end}):
\begin{align}
\mathbf{U}[u] \leftarrow \mathbf{U}[u] + g \cdot \mathbf{C}[v] - \eta \lambda \mathbf{U}[u] \\
\mathbf{C}[v] \leftarrow \mathbf{C}[v] + g \cdot \mathbf{U}[u] - \eta \lambda \mathbf{C}[v],
\end{align}
where $g$ is the signed gradient scalar for the sampled positive or negative interaction. When load imbalance is low (i.e., the standard deviation of local edge counts $E_p$, denoted $\sigma_{E_p}$, is small), larger batches increase the fraction of locally resolvable updates and reduce the relative communication imbalance.
% where $g$ denotes the signed gradient scalar derived from the sampled positive or negative interaction. If the load imbalance is low (i.e., standard deviation of $p's$ local edges $E_p$ is low; we denote this quantity as $\sigma_{E_p}$) with increasing batch sizes, the proportion of locally resolvable updates grows, reducing the relative imbalance between local computations and communication.

When a sampled pair involves a context vertex $v$ not owned by process $p$, updates are deferred rather than applied remotely or synchronized immediately, and computation proceeds on a locally available copy of the remote embedding. During batch training, $p$ buffers remote context vertices from positive and negative samples (Line~\ref{alg:defer-pos-remote},~\ref{alg:defer-neg-remote}). After local updates complete, processes issue \texttt{MPI\_Alltoallv} to exchange only the remote context embeddings needed by the current batch, keeping communication bounded (Line~\ref{alg:mpi-a2a-fetch}). Once available locally, the corresponding gradient updates are accumulated into a delta vector for each remote vertex (Line~\ref{alg:accumulate-delta-begin}--\ref{alg:accumulate-delta-end}):
% When a sampled pair involves a context vertex $v$ not owned by the current process, rather than performing immediate remote updates or synchronization, processes defer remote updates and operate on locally owned fetched copies of the remote embeddings. During batch training, $p$ buffers all the remote context vertices encountered in either positive or negative samples (Line~\ref{alg:defer-pos-remote},~\ref{alg:defer-neg-remote}). After completing local updates of the current batch, processes issue \texttt{MPI\_Alltoallv} to exchange the required remote context embeddings (Line~\ref{alg:mpi-a2a-fetch}). To keep communication volume within bounds, we only exchange the embeddings needed for the current batch. Once remote context embeddings are available locally,  processes compute their gradient updates using the fetched embeddings, accumulating them into a delta vector for each remote vertex (Line~\ref{alg:accumulate-delta-begin}--Line~\ref{alg:accumulate-delta-end}):
\begin{align}
\Delta[v] = \sum \left( g \cdot \mathbf{U}[u] - \eta \lambda \mathbf{C}[v] \right)
\end{align}
The sum aggregates all contributions involving $v$ within the batch, so $\Delta[v]$ is the net update to remote $C[v]$. After the batch, the accumulated $\Delta$ is exchanged via \texttt{MPI\_Alltoallv} (Line~\ref{alg:line:mpi-a2a-send-delta}), allowing each owner to apply the received deltas to its local context embeddings. Thus, all updates to a given embedding within a batch are applied by its owner.
% Above, the sum aggregates contributions from the updates involving $v$ within the current batch. $\Delta[v]$ represents the net update that should be applied to the remote $C[v]$. After processing the batch, the accumulated $\Delta$ is exchanged via another \texttt{MPI\_Alltoallv} (Line~\ref{alg:line:mpi-a2a-send-delta}), such that a remote process\slash owner can apply the received deltas into its local context embeddings. Thus, all modifications to a given embedding for a batch are applied by its owning processes.
% \note{write about algorithm 1}

MPI neighborhood collectives~\cite{hoefler2009sparse} could replace regular collectives for positive edges, since remote vertices reside on adjacent processes that share boundary nodes in the partitioned graph. However, for the randomly sampled negative edges, since they can be selected from non-adjacent processes, the existing communicator with the virtual process topology holding subsets of the partitioned graph would not suffice, and a new communicator must be created to which the resulting topology must be attached, which is an expensive operation (must be initiated every batch as needed). We attempted to use MPI neighborhood collective, choosing negative samples from the adjacent processes, but found the neighbor collective performance to be subpar and quality declined due to constraining the negative samples from process neighborhood.

\subsection{Random Sampling}\label{method-random-sampling}

Algorithm~\ref{random-walk-algorithm} briefly illustrates our MPI-based random walk implementation. We use an iterative design that continues until the global walk length reaches a specified limit. During initialization, each process prepares lists of send targets and receive sources; the \emph{enumerate} routine returns a process and its index from these lists (e.g., Line~\ref{alg:line:enumerate-lists}). For each source and target, user-defined buffers of size $SIZE$ store intermediate edges along the walk path. The walk starts from an arbitrary edge $\{u,v\}$, where $u$ is locally owned and $v$ may be local or remote, and inserts frontiers into queues for level-by-level exploration, similar to distributed Breadth First Search except that vertices may be revisited. Nonblocking receives are pre-posted (Line~\ref{alg:line:prepost-recvs}) to improve small-message performance (eager mode of MPI, common scenario since we periodically poke progress despite buffering).
% We briefly illustrate our MPI-based random walk implementation in Algorithm~\ref{random-walk-algorithm}. We adopt an iterative implementation which continues until the global walk length (or steps) reaches a specified limit. During initialization, every process prepares a list of processes it can send to (referred as \emph{targets}) and receive data from (referred as \emph{sources}). Inspired from Python, the \emph{enumerate} routine can fetch a tuple comprising of a neighborhood process and the corresponding index from the lists (e.g., see Line~\ref{alg:line:enumerate-lists}). For every process in \emph{sources} and \emph{targets}, user defined buffer sizes (depicted as $SIZE$ in Algorithm~\ref{random-walk-algorithm}) is maintained to send\slash receive the intermediate edges in the random walk path. The random walk begins from an arbitrary edge $\{u, v\}$ (assuming $u$ to be owned by the local process whereas $v$ could be either remote\slash local), with frontiers inserted into the queues for level by level exploration (similar to distributed Breadth First Search, except vertices can be revisited during random walks). Nonblocking receives are pre-posted as shown in Line~\ref{alg:line:prepost-recvs}, which can improve performance for small messages (eager mode of MPI, common scenario since we periodically poke progress despite buffering). 

Lines~\ref{alg:line:local-start}--\ref{alg:line:local-end} correspond to the local portion of the walk, where the next frontier is sampled from a discrete distribution over locally owned weighted edges. Posted nonblocking sends are tested (Lines~\ref{alg:line:test-posted-isends_start}--\ref{alg:line:test-posted-isends_end}), and completed buffers, counters, and request handles are reset. As the global step counter reaches the limit, a final round of zero-length messages is flushed to prepare for exit (Line~\ref{alg:line:prepare-exit}). Within the main loop, incoming requests are tested (starting at Line~\ref{alg:line:process-incoming-begin}) and inserted into the next-frontier queue; the current and next queues are swapped if the walk continues (Line~\ref{alg:line:swap-queues}).
% Lines~\ref{alg:line:local-start}--\ref{alg:line:local-end} correspond to the local portion of the random walk, where the next frontier is determined by sampling from a discrete distribution over locally owned weighted edges. The status of the outgoing nonblocking sends are tested  (Lines~\ref{alg:line:test-posted-isends_start}--\ref{alg:line:test-posted-isends_end}), and buffers\slash counters\slash MPI request handles associated with the completed send operations are reset. The final round of zero-length messages are flushed out as the global step counter reaches the limit (Line~\ref{alg:line:prepare-exit}), in preparation of exit. Within the main loop, incoming message requests are tested (beginning on Line~\ref{alg:line:process-incoming-begin}) and inserted into the next frontier queue; the current and next queues are swapped if the walk continues (Line~\ref{alg:line:swap-queues}).

%%%%%%%%%%%%%%%%%%%%%%%%%%%%%%%%%%%%%%%%%%%%%%
\begin{algorithm}[t]
{\scriptsize
\caption{ 
\small
Random Walks using MPI peer-to-peer interface.
\newline \textbf{Input}: $G_p=(V_p,E_p)$ partition of (undirected) graph $G$ in process $p$, containing ``local'' and ``ghost'' vertices (not owned by process $p$) that have an edge with a vertex $v \in V_p$. $SIZE$ is the outgoing\slash incoming buffer size, $steps$ is the walk length, \{u, v\} is the root edge to begin the walk and $\mathcal{N}(x)$ represents the (weighted) neighborhood of an arbitrary vertex $x$.
\newline \textbf{InOut}: Edges in random walk over $G$: $\mathcal{P}$ (local) and $\mathcal{P'}$ (remote).}
\label{random-walk-algorithm}
\begin{algorithmic}[1]
  \State \Comment{{\bf \textcolor{black}{--- Initialization ---}}}
  \State $sources, targets \leftarrow$ \{ \emph{list} of neighboring processes to \emph{send to}\slash \emph{receive from} \}
  \State $\{rbuf[0..|sources|\times SIZE] ,sbuf[0..|targets|\times SIZE]\} \leftarrow \emptyset$ \Comment{Buffers}
  \State $rreqs[0..|sources|] \leftarrow \emptyset,sreqs[0..|targets|] \leftarrow \emptyset$ \Comment{MPI request handles}
  \State $\forall x, t \in \mathtt{enumerate}(targets) \Rightarrow ok\_send[x] = false$ \label{alg:line:enumerate-lists} \Comment{Outgoing messages} 
  \State $\forall v \in V_p: visited[v] \leftarrow 0$ \Comment{tracks visited vertices}
  \Procedure{RandomWalk}{$G_p, \mathcal{P}, \mathcal{P'}, root=\{u, v\}$}
  \State $cq, nq \leftarrow \emptyset$ \Comment{queues for storing vertex frontiers}
  \State $curr\_steps \leftarrow 0$  \Comment{tracks global walk length}
  \State $\mathtt{visited[u]}\mathrel{+}=1 \mid \mathcal{P} \leftarrow \mathcal{P} \cup \{u, v\}$ \Comment{Assume vertex u is local}
  \If{$\mathtt{owner}(v) == p$} \Comment{Vertex v is local or remote}
  \State $\mathtt{visited[v]}\mathrel{+}=1 \mid cq \leftarrow cq \cup v$ \Comment{Initialize current queue}
  \Else
  \State $sbuf[\mathtt{owner}(v)] \leftarrow sbuf[\mathtt{owner}(v)] \cup \{u, v\}$
  \EndIf
  \For{$y, s \in \mathtt{enumerate}(sources)$} \Comment{Prepost incoming messages} \label{alg:line:prepost-recvs}
  \State $\mathtt{MPI\_Irecv}(rbuf[y], SIZE, s, rreqs[y])$
  \EndFor
  \While{$!recvs\_done \;\mathtt{and}\; curr\_steps < steps$} \label{alg:line:superstep-begin}
  \State \Comment{{\bf \textcolor{black}{--- Local random walk and post outgoing messages ---}}}
  \If{$curr\_steps \geq steps$} \Comment{prepare to exit} \label{alg:line:prepare-exit}
  \For{$x, t \in \mathtt{enumerate}(targets) \mid \; !msg\_sent[x]$}
  \State $\mathtt{MPI\_Isend}(NULL, 0, t, sreqs[t])$ \Comment{zero-byte sends mark exit}
  \State $ok\_send[x] \leftarrow true$
  \EndFor
  \Else \Comment{Poke communication progress and check current queue}
  \While{$cq \notin \emptyset$} \label{alg:line:local-start}
  \State $x: x \in cq \;\mathtt{and}\; cq \leftarrow cq \setminus x$ \Comment{pop from current queue}
  \State $\mathcal{X}: \mathcal{X} \sim discrete(\mathcal{N}(x)) \;\mathtt{and}\; nxt\_x \leftarrow \mathcal{X}(x)$ \Comment{distribution}
  \State $target, tidx \leftarrow owner(nxt\_x), targets[owner(nxt\_x)]$
  \If {$\mathit{target} == p$} \Comment{mark visited for local vertices} 
  \State $visited[nxt\_x] \mathrel{+}= 1 \mid nq \leftarrow nq \cup nxt\_x \mid \mathcal{P} \leftarrow \mathcal{P} \cup \{x, nxt\_x\}$ \label{alg:line:local-end}
  \Else \Comment{Initiate send if buffer is full, store otherwise}
  \If{$\mathtt{size}(sbuf[tidx]) == SIZE \; and \; !\mathit{ok\_send}[tidx]$} 
  \State $\mathtt{MPI\_Isend}(sbuf[tidx], SIZE, target, sreqs[tidx])$
  \State $ok\_send[tidx] \leftarrow true$
  \Else
  \State $sbuf[tidx] \leftarrow sbuf[tidx] \cup \{x, nxt\_x\}$
  \EndIf
  \EndIf
  \EndWhile
  \EndIf
  \State \Comment{{\bf \textcolor{black}{--- Process incoming data ---}}}
  \For{$y, s \in \mathtt{enumerate}(sources)$} \label{alg:line:process-incoming-begin}
  \If{$\mathtt{MPI\_Test}(rreqs[y])$} \Comment{Check incoming data message status}
  \State $count \leftarrow MPI\_Get\_count(s)$
  \If{$count == 0$} \Comment{zero-byte message received}
  \State $recvs\_done \mathrel{+}= 1$
  \Else \Comment{process receive buffer and append next frontier}
  \State $\forall \{m,n\} \in rbuf[y]:$
  \State $visited[n] \mathrel{+}= 1 \mid nq \leftarrow nq \cup \{m, n\} \mid \mathcal{P'} \leftarrow \mathcal{P'} \cup \{m, n\}$
  \State $\mathtt{MPI\_Irecv}(rbuf[y], SIZE, s, rreqs[y])$ \Comment{Prepost next}
  \EndIf
  \EndIf
  \EndFor
  \State \Comment{{\bf \textcolor{black}{--- Test completion ---}}}
  \State $sends\_done \leftarrow \mathtt{MPI\_Testsome}(|targets|, sreqs)$ \Comment{Test completion} \label{alg:line:test-posted-isends_start}
  \State{$\forall c \in sends\_done \Rightarrow sbuf[c] \leftarrow \emptyset, \; ok\_send[c] \leftarrow false$} \Comment{Clear} \label{alg:line:test-posted-isends_end}
  \State {$recvs\_done \leftarrow MPI\_Allreduce(recvs\_done)$} \Comment{check if done}
  \If{$recvs\_done < |sources|$}
  \State{$curr\_steps \leftarrow MPI\_Allreduce(\sum visited)$} \Comment{count visited}
  \If{$nq \neq \emptyset \; \mathtt{and} \; curr\_steps < steps$} \Comment{swap queues} \label{alg:line:swap-queues}
  \State $\mathtt{swap}(cq, nq)$
  \EndIf
  \EndIf
  \EndWhile
  \State \Return $|\mathcal{P}|/2$ \Comment{$\mathcal{P'}$ will be assembled later from remote processes}
  \EndProcedure
\end{algorithmic}
}
\end{algorithm}
%%%%%%%%%%%%%%%%%%%%%%%%%%%%%%%%%%%%%%%%%%%%%%
Algorithm~\ref{generate-pairs} repeatedly invokes random walks (Algorithm~\ref{random-walk-algorithm}) from roots sampled from a weight-normalized edge distribution until sufficient global samples (corresponding to the user-specified batch size) are collected for the current batch (Lines~\ref{alg:line:random-edge-selection}--\ref{alg:line:random-walk-algo}). This requires global synchronization to determine the exit condition (Line~\ref{alg:line:sampling-sync}), with two options: terminate when at least one process reaches the target batch size, or only when all processes do. We use \texttt{MPI\_Allreduce} for the former and a nonblocking MPI barrier for the latter (Lines~\ref{alg:line:ibarrier-sync-start}--\ref{alg:line:ibarrier-sync-end}).
% In Algorithm~\ref{generate-pairs}, we discuss the process of invoking random walks repeatedly (i.e., Algorithm~\ref{random-walk-algorithm}) from roots selected from a discrete probability distribution of edges (normalized by weights) until sufficient global samples (corresponding to the user specified batch size) are collected for the current batch (Lines~\ref{alg:line:random-edge-selection}--\ref{alg:line:random-walk-algo}). This necessitates a global synchronization to determine the exit criteria (Line~\ref{alg:line:sampling-sync}), and there are two choices: exit when at least one of the processes have reached the designated batch size or exit when \emph{all} the processes have done so. We use \texttt{MPI\_Allreduce} based synchronization for the former, and utilize MPI nonblocking barrier for the latter (Lines~\ref{alg:line:ibarrier-sync-start}--\ref{alg:line:ibarrier-sync-end}).
Each process invokes the nonblocking barrier after collecting the required number of batches; completion is detected when all processes in the communicator have entered the barrier and the associated MPI request handle returned true. This synchronization choice affects per-process working sets and can either worsen or ameliorate load imbalance and scalability, depending on the input. After random-walk generation completes, the remote portion of the collected samples is assembled via \texttt{MPI\_Alltoallv} (after exiting the loop between Lines~\ref{alg:line:begin-loop}--\ref{alg:line:end-loop}), as shown in Line~\ref{alg:line:assemble-remote-samples} (deliberately excluded from Algorithm~\ref{random-walk-algorithm} to reduce the communication overhead).
% Each process invokes the nonblocking barrier after the specified number of batches were collected. When all the processes in the MPI communicator have invoked the barrier, testing the MPI request handle associated with it shall return true, indicating completion. The choice of synchronization affects the working set of individual processes, which can worsen or ameliorate the load imbalance and scalability for particular inputs. The remote portion of all the samples collected must be assembled via another \texttt{MPI\_Alltoallv} after the random walks are done (i.e., after exiting the loop between Lines~\ref{alg:line:begin-loop}--\ref{alg:line:end-loop}), as shown in Line~\ref{alg:line:assemble-remote-samples} (deliberately excluded from Algorithm~\ref{random-walk-algorithm} to reduce the communication overhead). 

%%%%%%%%%%%%%%%%%%%%%%%%%%%%%%%%%%%%%%%%%%%%%%
\begin{algorithm}[t]
{\scriptsize
\caption{ 
\scriptsize
Invoking random walks for generating batches for sampling.
\newline \textbf{Input}: $G_p=(V_p,E_p)$ portion of the (undirected) graph $G$ on process $p$.  
\newline \textbf{InOut}: Edges in random walk over $G$: $\mathcal{P}$ (local) and $\mathcal{P'}$ (remote).}
\label{generate-pairs}
\begin{algorithmic}[1]
\Procedure{GeneratePairs}{$G_p, B, \mathtt{sync\_type}=allreduce$}
\State $pairs \leftarrow 0 \mid done \leftarrow false \mid bar \leftarrow false$
\State $nb\_req \leftarrow MPI\_REQUEST\_NULL$ \Comment{nonblocking barrier request}
\While{$!done$} \label{alg:line:begin-loop}
\State \Comment{{\bf \textcolor{black}{--- Random Walk ---}}}
\State $\{u, v\} \leftarrow \mathcal{X}(): \mathcal{X} \sim discrete(\{\vec{E_p} / \left\lVert \vec{E_p} \right\rVert\})$ \Comment{random selection} \label{alg:line:random-edge-selection}
\State $pairs \mathrel{+}= \textsc{RandomWalk}(G_p, \mathcal{P}, \mathcal{P'}, root=\{u, v\})$ \label{alg:line:random-walk-algo}
\State \Comment{{\bf \textcolor{black}{--- Synchronization (Allreduce or Ibarrier) ---}}} \label{alg:line:sampling-sync}
\If{$\mathtt{sync\_type} == allreduce$} \Comment{allreduce based synchronization}
\State $pairs \leftarrow MPI\_Allreduce(|\mathcal{P}|, \mathtt{default\_op}=\mathtt{MAX})$  \label{alg:line:allreduce-sync}
\If{$pairs \geq B$} \Comment{true $\mathit{iff}$ any process collect $samples \geq B$}
\State $done \leftarrow true$ 
\EndIf
\Else \Comment{nonblocking barrier based synchronization} \label{alg:line:ibarrier-sync-start}
\If{bar} \Comment{nonblocking barrier is activated}
\If{$MPI\_Test(nb\_req)$} \Comment{test for completion}
\State $done \leftarrow true$
\Else
\If{$pairs \geq B$} \Comment{true $\mathit{iff}$ all processes collect $samples \geq B$}
\State $MPI\_Ibarrier(nb\_req)$ \Comment{activate nonblocking barrier}
\State $bar \leftarrow true$
\EndIf 
\EndIf \label{alg:line:ibarrier-sync-end}
\EndIf
\EndIf
\EndWhile \label{alg:line:end-loop}
\State $\mathcal{P} \leftarrow MPI\_Alltoallv(\mathcal{P'})$ \Comment{assemble walk paths from remote} \label{alg:line:assemble-remote-samples}
\State \Return $\shuffle_2\mathcal{P} \rightarrow \mathcal{P}_{1:B}$ \Comment{Shuffle by pairs and prune to batch size}\label{alg:line:prune-batch-before-return}
\EndProcedure
\end{algorithmic}
}
\end{algorithm}
%%%%%%%%%%%%%%%%%%%%%%%%%%%%%%%%%%%%%%%%%%%%%%

%\subsection{Remote Updates}

%\subsection{Communication}

% talk about all the remote variants and why
\subsection{Variants of random walks} \label{method-remote-variants}
Sampling drives \nomad's performance at scale, so we propose several trade-offs to mitigate the communication costs induced by irregular graph structure. Under the SPMD model, all processes participate in sampling and communication, making design choices largely dependent on the resulting load imbalance and synchronization overhead during repeated random walks. Following Algorithms~\ref{alg:nomad_simplified}, \ref{random-walk-algorithm}, and \ref{generate-pairs}, the overall communication cost of \nomad can be represented by Eqn.~\ref{eqn:cost_nomad}, which spans both sampling and training over all batches.
% Sampling drives the performance of \nomad at scale, and we propose several trade-offs to mitigate the communication, which can become severe due to the inherently irregular structure of the input graphs. Due to the SPMD nature of the problem, all processes must participate in sampling and communication.  As such, our choices are driven by the load imbalance and associated synchronization\slash communication underwent during repeated invocations of random walk. Following Algorithms~\ref{alg:nomad_simplified}, ~\ref{random-walk-algorithm} and ~\ref{generate-pairs}, the overall communication overhead of \nomad can be represented by Eqn.~\ref{eqn:cost_nomad}, comprising of the random sampling and training phases for the duration of the number of batches.   
\begin{align}\label{eqn:cost_nomad}
{\small
(\underbrace{(\underbrace{\log p}_{\text{synch}} + \underbrace{p\times SIZE}_{\text{point-to-point}})}_{\text{until $\; |\mathcal{P}| = \mathbb{B}\;$ for any\slash all $\;p$}} + \underbrace{p \times |\mathcal{P}_{1:\mathbb{B}}|}_{\text{all-to-all-v at training}}) \times \#\mathtt{batches}
}
\end{align} 
Eqn.~\ref{eqn:cost_nomad} shows that in distributed memory, random sampling is the scalability bottleneck because training cannot proceed until samples are generated. For irregular traversals, communication buffers are kept small to limit load imbalance by frequently ensuring communication progress ($SIZE=128$ in \nomad). Reducing communication costs during sampling is critical. As shown in Algorithm~\ref{generate-pairs}, we therefore consider both nonblocking-barrier and all-reduce-based synchronization, along with variants that reduce the number of distributed walks by augmenting vertex pairs, reusing pairs from prior walks (current samples are pruned to the batch size; Line~\ref{alg:line:prune-batch-before-return}), or generating samples upfront outside the training loop (before Line~\ref{alg:line:before-training-loop} in Algorithm~\ref{alg:nomad_simplified}). Since load imbalance can arise either from partitioning or from insufficient samples on a process (depending on the synchronization strategy (Algorithm~\ref{generate-pairs}), with the latter often more detrimental because communication dominates, distributed random-walk sampling presents trade-offs among \emph{sample quality}, \emph{communication\slash synchronization cost}, and \emph{memory requirement}. We explore this design space through six \nomad random-walk variants, each representing a distinct tradeoff point.

\begin{enumerate}[left=0pt, nosep]
\item \emph{\textbf{Local}}: Each process walks only over locally owned vertices, terminating when the next vertex is remote (Lines~\ref{alg:line:local-start}--\ref{alg:line:local-end}, Algorithm~\ref{random-walk-algorithm}). This is the most scalable variant (avoids communication), but limited exploration may require multiple walk invocations from different roots to generate sufficient training samples, potentially reducing quality.
 % \item \emph{Local}: Every process performs walk only traversing locally owned vertices, exiting when the next vertex on the path is remote; this is shown in Lines~\ref{alg:line:local-start}--\ref{alg:line:local-end} of Algorithm~\ref{random-walk-algorithm}. Since this version is communication free, it is the most scalable. However, it may require several invocations of random walk (with different roots) to generate sufficient training samples, offering limited exploration, affecting the quality. 

 \item \emph{\textbf{Remote (Fresh-Only)}}: This variant is invoked within the training loop (e.g., Line~\ref{alg:line:pos-sample} of Algorithm~\ref{alg:nomad_simplified}), so each batch is built from newly generated remote walks, prioritizing sample quality over reuse. However, invoking distributed remote walks hundreds to thousands of times per batch introduces substantial communication overhead and high susceptibility to load imbalance.
 % \item \emph{Remote (Fresh-Only)}: This variant is invoked within the training loop (like Line~\ref{alg:line:pos-sample} in Algorithm~\ref{alg:nomad_simplified}), such that each batch is ``freshly'' constructed from the newly generated remote random walks, prioritizing sample quality (as opposed to reusing outdated samples). However, this variant is susceptible to nontrivial communication overheads and load imbalances, as distributed remote walk is invoked hundreds--thousands of times in every batch.
\item \emph{\textbf{Remote (Reuse-Spill)}}: To amortize random-sample generation, this variant retains unused samples from prior batches in a persistent ``spill" buffer, at the cost of extra memory. Instead of discarding samples beyond the batch limit as in Remote (Fresh-Only) (Line~\ref{alg:line:prune-batch-before-return}, Algorithm~\ref{generate-pairs}), it reuses them in later batches before invoking additional remote walks, reducing communication at scale. However, quality can degrade when later batches rely primarily on stale spill-buffer samples rather than fresh walks.
 % \item \emph{\textbf{Remote (Reuse-Spill)}}: To amortize random-sample generation, this variant retains unused samples from prior batches in a persistent ``spill" buffer, at the cost of extra memory. Rather than discarding samples beyond the batch limit as in Remote (Fresh-Only) (Line~\ref{alg:line:prune-batch-before-return} of Algorithm~\ref{generate-pairs}), they are reused in subsequent batches before invoking additional remote walks, reducing communication at large scale. However, quality can degrade when many later batches are trained primarily from stale spill-buffer samples rather than fresh walks.
 % \item \emph{Remote (Reuse-Spill)}: To amortize the cost of generating random samples, the reuse-spill variant retains unused samples across batches in a persistent ``spill'' buffer at the expense of extra memory. Contrastingly, in the fresh-only version, as shown in Line~\ref{alg:line:prune-batch-before-return} of Algorithm~\ref{generate-pairs}, generated samples are pruned to the batch size; instead of throwing the extra samples, those are pushed into the spill buffer for use in subsequent batches. In each batch, samples are first drawn from the spill buffer instead of invoking additional remote walks. This approach can reduce overall communication, particularly of relevance at large scale. The sample quality can be compromised in this version, as there can be a scenario where a significant number of the remaining batches are trained solely with  samples from the persistent\slash spill buffer, not requiring any fresh samples, biasing the training.

 \item \emph{\textbf{Remote (Refresh-Spill)}}: Like reuse-spill, this variant uses a spill buffer while invoking random walks within the training loop, but spilled samples are discarded after one time use (only from previous batch) rather than retained indefinitely. This mitigates sample bias while still amortizing communication across batches. Relative to reuse-spill, it improves sample quality at the cost of higher synchronization overhead.
 % \item \emph{Remote (Refresh-Spill)}: This variant also employs a spill buffer like the aforementioned reuse-spill variant while invoking random walks within the training loop. But, rather than indefinitely retaining and reusing samples, it discards them once used (essentially it only uses the spilled pairs from the previous batch). This strategy mitigates sample bias while amortizing communication costs across multiple batches. Compared to reuse-spill, this variant improves sample quality at the expense of higher synchronization overhead.

 \item \emph{\textbf{Remote (Augmented-Single)}}: This variant performs a single long walk from one root, with walk length scaled to batchsize$\times$\#batches, before training begins. It is the slowest variant because samples for all batches must be generated upfront. The resulting samples are also augmented from a single traversal root, which can degrade quality.
 % \item \emph{Remote (Augmented-Single)}: This variant performs a single long walk (adjusting the walk size or steps to be proportional to batch size $\times$ \#batches) from a single root, instead of using different roots for several rounds of random walks, before training is initiated. Consequently, this is the slowest variant because samples for all the batches must be collected in advance before training can resume. Also, edges are augmented in the resultant samples due to walk from a distinct root, affecting the overall quality.
 \item \emph{\textbf{Remote (Augmented-Pair)}}: This variant also decouples sampling from training by generating samples upfront, but invokes remote walks only until roughly $\sqrt{batch\;size \times \#batches \times 2}$ pairs are collected, then expands them by enumerating additional pair combinations. Unlike Augmented-Single, whose upfront samples are always positive, Augmented-Pair can introduce negative pairs through recombination. Its goal is to reduce load imbalance (no variation in min\slash max samples as in the non-augmented remote variants) by giving each process the same number of samples (middle ground between Local and rest of the Remote variants). Both augmented variants use \texttt{MPI\_Ibarrier}-based synchronization (see \S\ref{method-random-sampling}) to ensure sufficient sample availability.

% \item \emph{Remote (Augmented-Pair)}: The augmented-allpair variant also generates a sequence of node pairs in advance by continuously invoking remote walks before training is initiated (i.e., decoupling random sampling from training). However, the idea is to generate pairs by invoking random walks until a factor equivalent to the square root of $batch\; size \times \#batches \times 2$ is reached. Then, the generated pairs are augmented by enumerating different pairs at a time, expanding the original samples. While both Augmented-Single\slash Pair variants prepares the samples in advance rather than during training, the pairs generated by Augmented-Single are always positive, whereas there can be negative pairs due to combinations in Augmented-Pair. The intention of this version is to provide a middle ground between Local and rest of the Remote variants, offering less load imbalance since each process has the same amount of samples (no variation in min\slash max samples as in the non-augmented remote variants). To ensure sufficient sample availability, both the augmented versions utilize \texttt{MPI\_Ibarrier} based synchronization (refer to \S\ref{method-random-sampling}).
\end{enumerate}

\section{Results}\label{sec:results}
We evaluate \nomad on real-world graphs with varied parameterizations (\S\ref{results-setup}), evaluate baseline comparisons (\S\ref{ssec:baseline}), discuss quality/performance tradeoffs relative to state-of-practice (\S\ref{ssec:results-quality}), and analyze performance trade-offs(\S\ref{ssec:results-performance}).
%distributed-memory 
\subsection{Experimental Setup}\label{results-setup}
% For checking raw numbers
% \midrule
% pubmed               & 20K  & 89K   & 12$\slash$152 & 12$\times$ \\
% photo                & 7.5K & 238K  & 16$\slash$1091 & 37$\times$\\
% computers            & 13K  & 491K  & 39$\slash$3193 & 58$\times$\\
% physics              & 34K  & 495K  & 9$\slash$181 & 23$\times$ \\
% ogbn-arxiv           & 169K & 2.3M  & 60$\slash$2456 & 99$\times$\\
% youtube              & 1.1M & 9.8M   & 199$\slash$5788 & 370$\times$  \\
% ogbn-proteins        & 132K & 79M   & 1802$\slash$13427 & 266$\times$  \\
% reddit               & 233K & 114M  & 1449$\slash$34900 & 167$\times$ \\
% ogbn-products        & 2.4M & 123.7M & 172$\slash$21477 & 315$\times$  \\
% \midrule
% hyperlink2012        & 39M   & 1.1B  & 376596$\slash$104031 & 22$\times$  \\
% twitter7             & 41.6M & 2.9B  & $1.21 \times 10^{6}$$\slash$346191 & 14$\times$ \\
% friendster           & 65.6M & 3.6B  & 6395$\slash$396386 & 18$\times$\\
% ogbn-papers100M      & 111M  & 6.4B  & 13859$\slash$225013 & 19$\times$\\
% \bottomrule

\subsubsection{Datasets/Platform/Variants}\label{results-setup-datasets}
We evaluate \nomad on diverse real-world graph datasets spanning citation, social, and web networks (Table~\ref{tab:datasets}); several are accessed through DGL's dataset APIs, while the original sources are listed in the table. We convert the graphs from native format to an undirected binary CSR representation for efficient I/O. To characterize partition irregularity (see \S\ref{sec:methods}), we report the standard deviation of edges across processes, $\sigma_{E_p}$, since higher values can worsen load imbalance. Table~\ref{tab:datasets} also highlights the best observed speedup for each input across process configurations and variants. Experiments are performed on the NERSC Perlmutter consisting of 3,072 CPU-only compute nodes. Each node uses two 64-core AMD EPYC 7763 CPUs (128 cores / 256 threads) and 512 GB DDR4 memory, eight NUMA domains, and is connected via Slingshot 11~\cite{yang2020accelerate}. We use up to 512 nodes in our experiments. \nomad is built with PrgEnv-gnu/8.5.0 (GNU v13.2), cray-libsci/24.07.0 (for BLAS \texttt{sdot}) and cray-mpich/8.1.30 for building \nomad, and CrayPAT\slash Perftools (perftools-base/24.07.0) for profiling.

Experimental variants of \nomad are denoted in this format \textless \textbf{Sync}:AR\slash\textsc{IBarrier}(IB)\textgreater--\textless\textbf{Variant}:Local\slash Remote(Fresh\slash Reuse-Spill\slash Refresh-Spill\slash Aug-Single\slash Aug-Pair)\textgreater (augmented variants use \textsc{IBarrier} only), following \S\ref{method-random-sampling} and \S\ref{method-remote-variants}.
\subsubsection{Baselines}\label{results-baselines}
We use different baselines for quality and performance comparisons. For embedding quality, we compare against the original LINE implementation~\cite{LINE_github}, an optimized C++ implementation of node2vec~\cite{snap_github}, and hybrid CPU--GPU-based GraphVite~\cite{zhu2019graphvite} (hyperparameters were matched to \nomad as closely as permitted by design). For performance, we compare \nomad against multithreaded LINE and node2vec in the shared-memory setting (128 CPU threads; max. on Perlmutter), and against PyTorch-BigGraph (PBG)~\cite{lerer2019pytorch} in the distributed-memory setting (using recommended \#partitions=$2\times$\#nodes; a 128 partitions/node variant, evaluated for closer alignment with \nomad, was about 18$\times$ slower). GraphVite is excluded as it is directly not comparable to \nomad in end-to-end distributed-memory performance (See \S\ref{sec:intro}).
% GraphVite is included only in the quality analysis (hyperparameters were matched to \nomad as closely as permitted by its interface). since it is a single-node hybrid CPU--GPU system and therefore not directly comparable to \nomad in end-to-end distributed-memory performance.
% LINE and node2vec use shared-memory multithreading on a single Perlmutter node (128 CPU threads), whereas \nomad distributes work across MPI ranks. 
Following prior work~\cite{zhu2019graphvite,tang2015line}, we evaluate embedding quality using downstream multi-class node classification with a one-vs-rest logistic regression model trained on 10\% labeled nodes, and report Micro-F1 and Macro-F1. 

\subsubsection{\nomad Hyperparameters}\label{results-nomad-params}
%NOMAD trains $d$-dimensional node embeddings, where $d$ is set via the embedding dimension parameter. Optimization is performed using stochastic gradient descent with learning rate $\eta$ (default $\eta{=}0.025$). For each positive context pair generated from random walks, NOMAD samples $K$ negative examples (we use $K{=}1$). Training proceeds in batches of fixed size, which determines the number of positive samples processed per training iteration (we use batchsize $100-100K$ based on the size of the graph). The total training budget is controlled by the number of epochs, where each epoch corresponds to a full pass over the walk generated samples. Random-walk sampling is controlled by the walk length parameter (we use $100$) and the window size (we use $2$), which together determine the number and locality of positive pairs extracted from each walk. NOMAD uses $\ell_2$ regularization (weight decay) to improve generalization.
Following standard settings, \nomad trains $d$-dimensional embeddings using single-precision Stochastic Gradient Descent (SGD) with learning rate $\eta$ (default $0.025$), one negative sample per positive pair ($K{=}1$), and $\ell_2$ regularization. Training uses fixed-size batches ($100$--$100$K, depending on the input), with epochs controlling the overall budget. Random-walk sampling is parameterized by walk length (typically $100$) and window size (typically $2$), which determine the number and locality of positive pairs. 
Table~\ref{tab:datasets} summarizes the main trends. \nomad achieves substantial end-to-end speedups across diverse graphs ($12\times$--$58\times$ on small and $167\times$--$370\times$ on medium-to-large), but scalability is governed primarily by positive-pair generation rather than embedding updates (\S\ref{ssec:results-scalability}). 
\begin{table}[t]
\centering
\small
\caption{\small Best speedups observed across processes\slash variants.}
\label{tab:datasets}
\resizebox{\columnwidth}{!}{%
\begin{tabular}{|lrrrr|}
\hline
\textbf{Datasets} & \textbf{$|V|$} & \textbf{$|E|$} & \textbf{$\min/\max(\sigma_{E_p})$} & \textbf{Speedup} \\
\hline
\hline
pubmed~\cite{dgl_pubmed_dataset}               & 20K  & 89K   & 12$\slash$152 & 12$\times$ \\
photo~\cite{gnnbenchmark_datasets}                & 7.5K & 238K  & 16$\slash$1K & 37$\times$\\
computers~\cite{gnnbenchmark_datasets}            & 13K  & 491K  & 39$\slash$3K & 58$\times$\\
physics~\cite{gnnbenchmark_datasets}              & 34K  & 495K  & 9$\slash$181 & 23$\times$ \\
ogbn-arxiv\cite{hu2020open}           & 169K & 2.3M  & 60$\slash$2K & 99$\times$\\
youtube~\cite{mpisws_social_datasets}              & 1.1M & 9.8M   & 199$\slash$6K & 370$\times$  \\
ogbn-proteins\cite{hu2020open}        & 132K & 79M   & 2K$\slash$13K & 266$\times$  \\
reddit~\cite{snapnets}               & 233K & 114M  & 1K$\slash$35K & 167$\times$ \\
ogbn-products\cite{hu2020open}        & 2.4M & 123.7M & 172$\slash$21K & 315$\times$  \\
\hline
hyperlink2012~\cite{webdatacommons_hyperlink2012}        & 39M   & 1.1B  & 104K$\slash$376K & 22$\times$  \\
twitter7~\cite{snapnets}             & 41.6M & 2.9B  & 35K$\slash$1.21M & 14$\times$ \\
friendster~\cite{snapnets}           & 65.6M & 3.6B  & 6K$\slash$396K & 18$\times$\\
ogbn-papers100M\cite{hu2020open}       & 111M  & 6.4B  & 14K$\slash$225K & 19$\times$\\
\hline
\end{tabular}%
}
\end{table}

\subsection{Baseline Comparison}\label{ssec:baseline}
{
\setlength{\fboxsep}{3pt}
\noindent\fbox{%
\parbox{0.97\columnwidth}{%
\nomad provides 4-291$\times$ speedup over single-node baselines LINE and node2vec, and 35-76$\times$ speedup over distributed-memory baseline PyTorch-BigGraph (PBG). 
%across small\slash medium-scale graphs.
%
}}
}

\begin{figure}[t]
  \centering
  \includegraphics[width=\linewidth]{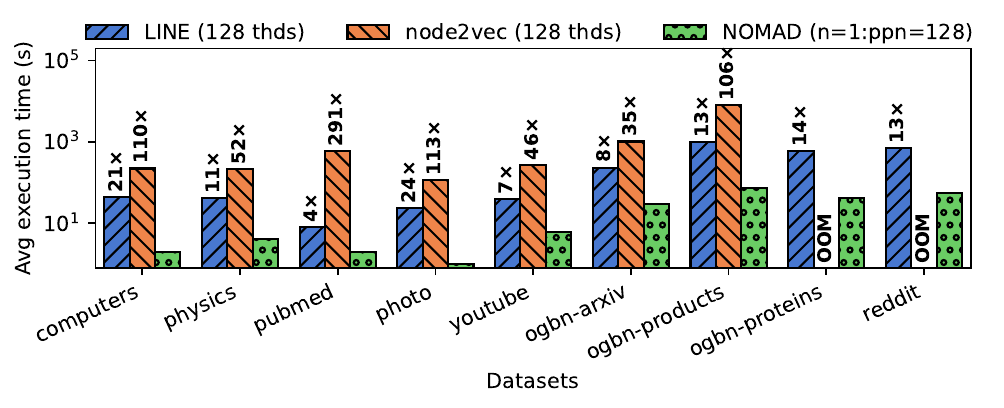}
  % \vspace{-2em}
  \caption{Comparison of \nomad with LINE and node2vec.}
  \label{fig:baseline-speedup}
\end{figure}

For the single-node comparison, we evaluate \nomad on a single Perlmutter node (n=1:ppn=128) against multithreaded LINE and node2vec, which are limited to a single node with fixed memory (cannot accommodate medium-large graphs). Fig.~\ref{fig:baseline-speedup} shows, \nomad maintains substantially lower end-to-end execution time. 

\begin{figure}[t]
  \centering
  \includegraphics[width=\linewidth]{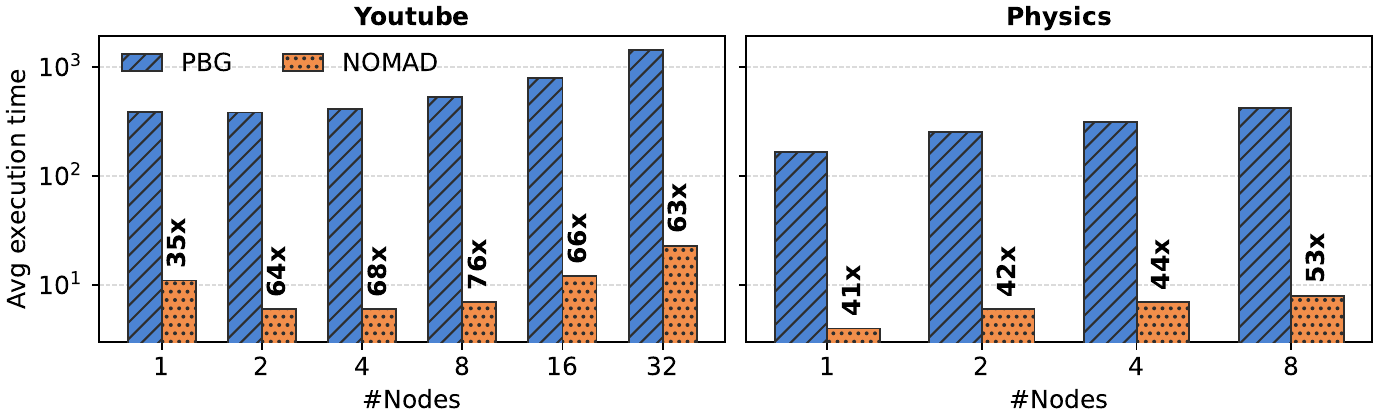}
  \caption{Comparison of \nomad with PBG}
  \label{fig:baseline-dist-speedup}
\end{figure}
% Across datasets as shown in Fig.~\ref{fig:baseline-speedup}, \nomad (n=1:ppn=128) achieves a median speedup range of 13--52$\times$ over multithreaded LINE and node2vec. The scalability of multi-threaded LINE and node2vec is limited to a single node with fixed memory (cannot accommodate medium-large graphs).
% \subsubsection{Distributed-Memory}
Against PBG (distributed-memory), as shown in Fig.~\ref{fig:baseline-dist-speedup}, \nomad consistently achieves lower end-to-end execution time (41--53$\times$ speedup on physics and 35--66$\times$ on youtube). PBG incurs substantial overhead from its edge-bucket-based partition training model, which relies on lock-based bucket scheduling, repeated swapping of partitioned embeddings, and lazy distributed parameter synchronization. These costs become pronounced when runtime is not dominated by training alone (also observed by prior work~\cite{fang2023distributed,recht2011hogwild,renz2022nups}) and increasingly expensive on very large graphs. Hence, we restrict this comparison to only medium-scale graphs.
\begin{table}[b]
\centering
\small
\caption{\small Micro-F1 (Mi), Macro-F1 (Ma) on 10\% test nodes.}
\label{tab:f1_comparison}
\resizebox{\columnwidth}{!}{%
\begin{tabular}{lcc cc cc cc cc}
\toprule
\textbf{Dataset} &
\multicolumn{2}{c}{\textbf{LINE}} &
\multicolumn{2}{c}{\textbf{node2vec}} &
\multicolumn{2}{c}{\textbf{GraphVite}} &
\multicolumn{2}{c}{\textbf{\nomad (local)}} &
\multicolumn{2}{c}{\textbf{\nomad (remote)}} \\
\cmidrule(lr){2-3}\cmidrule(lr){4-5}\cmidrule(lr){6-7}\cmidrule(lr){8-9}\cmidrule(lr){10-11}
& \textbf{Mi} & \textbf{Ma} & \textbf{Mi} & \textbf{Ma} & \textbf{Mi} & \textbf{Ma} & \textbf{Mi} & \textbf{Ma} & \textbf{Mi} & \textbf{Ma} \\
\midrule
computers      & 0.61 & 0.50 & 0.50 & 0.41 & 0.79 & 0.74 & 0.83 & 0.82 & \textbf{0.83} & \textbf{0.83} \\
physics        & 0.59 & 0.31 & 0.74 & 0.69 & 0.60 & 0.41 & 0.78 & 0.69 & \textbf{0.83} & \textbf{0.77} \\
pubmed         & 0.41 & 0.30 & 0.48 & 0.43 & 0.40 & 0.22 & \textbf{0.62} & \textbf{0.53} & 0.53 & 0.40 \\
photo          & 0.77 & 0.73 & 0.51 & 0.48 & 0.77 & 0.70 & 0.84 & 0.82 & \textbf{0.89} & \textbf{0.88} \\
youtube        & 0.25 & 0.06 & \textbf{0.40} & \textbf{0.29} & 0.31 & 0.14 & 0.27 & 0.11 & 0.25 & 0.07 \\
ogbn-arxiv     & 0.31 & 0.04 & 0.45 & 0.22 & 0.44 & 0.10 & \textbf{0.54} & \textbf{0.21} & 0.26 & 0.05 \\
ogbn-products  & 0.28 & 0.02 & 0.55 & 0.23 & 0.59 & 0.24 & \textbf{0.72} & \textbf{0.31} & 0.71 & 0.31 \\
ogbn-proteins  & 0.62 & 0.48 & OOM & OOM & 0.62 & 0.46 & \textbf{0.66} & \textbf{0.54} & \textbf{0.66} & \textbf{0.54} \\
reddit         & 0.82 & 0.76 & OOM & OOM & 0.89 & 0.80 & \textbf{0.92} & \textbf{0.88} & 0.91 & 0.83 \\
\bottomrule
\end{tabular}%
}
\end{table}
\subsection{Embedding Quality}\label{ssec:results-quality}
{
\setlength{\fboxsep}{3pt}
\noindent\fbox{%
\parbox{0.97\columnwidth}{%
\nomad improves embedding quality by up to 29\% over the strongest single-node baseline across datasets (up to 157\%, 31\%, and 55\% over LINE, node2vec, and GraphVite). Best quality comes from variants that preserve sample integrity, while the fastest variants trade quality for execution time.
}}
}

% \subsection{Quality}\label{ssec:results-quality}
% We evaluate the quality of the embeddings generated by \nomad against LINE and node2vec, following \S\ref{results-baselines}.
%Table \ref{tab:f1_comparison} reports Micro-F1 and Macro-F1 scores on node classification using 10\% labeled nodes, where each method is evaluated using its best-performing configuration across all tested NOMAD at scale. Overall, NOMAD consistently achieves higher embedding quality than both LINE and node2vec across most datasets, demonstrating that distributed execution does not compromise representation quality. On small- and medium-scale graphs such as computers, physics, photo, and pubmed, NOMAD substantially improves over LINE and node2vec, with gains of up to 22–38 percentage points in Micro-F1. In particular, on computers and photo, NOMAD achieves near-perfect Macro-F1 scores, indicating strong class-level balance that is not captured by prior methods. On larger, more structurally complex graphs, including ogbn-products and reddit, NOMAD continues to outperform single-node baselines, achieving up to 44 percentage points higher Micro-F1 than LINE and remaining competitive with or exceeding node2vec where results are available. Notably, node2vec fails to complete on ogbn-proteins and reddit due to memory limit, while NOMAD successfully trains and maintains strong classification performance, highlighting its robustness at scale. LINE similarly exhibits degraded Macro-F1 on ogbn-arxiv and ogbn-products, suggesting limited expressiveness under high-degree and label-imbalanced settings.

Table~\ref{tab:f1_comparison} reports Micro-F1 and Macro-F1 for node classification using 10\% labeled nodes, considering the best-performing local and remote configurations of \nomad (\S\ref{method-remote-variants}). Overall, \nomad achieves higher or competitive quality relative to LINE, node2vec, and GraphVite across most datasets, showing that distributed execution does not compromise representation quality. On computers, physics, photo, and pubmed (small\slash medium-scale), \nomad improves Micro-F1 by 5--29\% over the strongest single-node baseline; on larger graphs such as ogbn-products and reddit, the improvement ranges from 3--22\%. On computers and photo, \nomad also achieves near-perfect Macro-F1, indicating strong class-level balance. For larger inputs such as ogbn-proteins and reddit, node2vec fails due to node-memory limitations, whereas \nomad maintains strong performance. LINE similarly exhibits degraded Macro-F1 on ogbn-arxiv and ogbn-products, suggesting limited expressiveness under high-degree and label-imbalanced settings. Overall, both local and remote \nomad variants provide strong quality guarantees on 7/9 inputs in Table~\ref{tab:f1_comparison}, likely because distributed-memory parallelism enables substantially more walk-derived training samples within the same experimental setting. On youtube, increasing the training budget from 10 to 100 epochs improves \nomad to 0.34/0.26 and 0.35/0.26 (Mi/Ma) for the best local and remote variants, respectively; we did not perform dataset-specific hyperparameter tuning.

\begin{figure}[t]
  \centering
  \includegraphics[width=\linewidth,trim={0 12 0 0},clip]{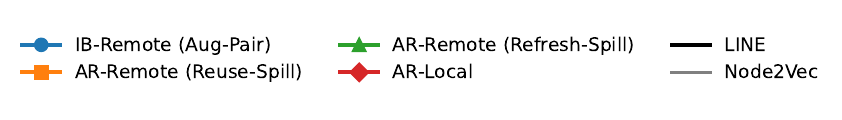}
  % \vspace{-1em}
  \begin{subfigure}[t]{0.49\linewidth}
    \centering
    \includegraphics[width=\linewidth]{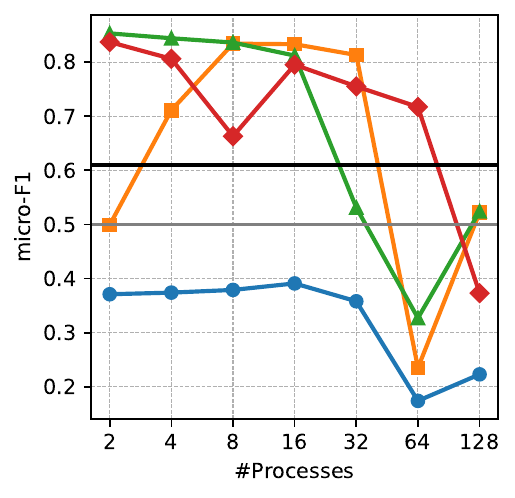}
    \label{fig:quality-computers}
  \end{subfigure}
  \hfill
  \begin{subfigure}[t]{0.49\linewidth}
    \centering
    \includegraphics[width=\linewidth]{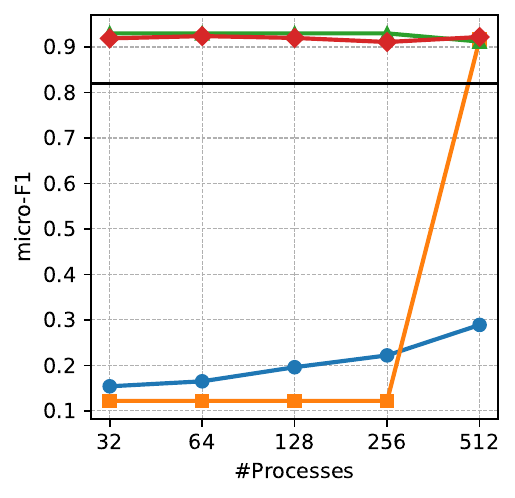}
    \label{fig:quality-reddit}
  \end{subfigure}
  \caption{Embedding quality (Micro-F1, higher is better) on computers (left) and reddit (right) with increasing \#processes.}
  \label{fig:quality}
\end{figure}

\subsubsection{Performance-Quality Tradeoff}\label{ssec:results-quality-tradeoff}
Fig.~\ref{fig:quality} demonstrates the quality of generated embeddings (using Micro-F1) of different variants of \nomad with increasing processes\slash parallelism on two diverse inputs: computers and reddit (depicts major variations in $\sigma_{E_p}$, see Table~\ref{tab:datasets}). 
Local sampling (\texttt{AR-Local}) consistently achieves high Micro-F1 across the process counts, indicating that locally generated random walks preserve the structural fidelity required for high-quality embeddings. However, as distribution increases, individual partition sizes may vary, abruptly affecting the sample quality (e.g., see computers between 4--8 and 64--128 processes in Fig.~\ref{fig:quality}). Remote execution strategies also exhibit variations on computers (see \texttt{AR-Remote} trajectories). However, remote variants that generates fresh samples (\texttt{AR-Remote (Refresh\slash Reuse-Spill)}) demonstrate competitive qualities to the locally generated samples. As discussed in \S\ref{method-remote-variants}, augmented random walks (i.e., \texttt{IB-Remote (Augmented-Pair)}) can lead to subpar quality (especially for small\slash medium inputs) due to overabundance of negative edges in samples, with no recourse to periodically refresh (consequently, it demonstrates the least Micro-F1). Macro-F1 trends are similar.

\subsection{Performance Analysis}\label{ssec:results-performance}
{
\setlength{\fboxsep}{3pt}
\noindent\fbox{%
\parbox{0.97\columnwidth}{%
%\nomad's variant 
\texttt{AR-Local} provides best performance/quality tradeoff. \texttt{AR-Remote} variants deliver best embedding quality at the expense of remote communication, without necessarily affecting the performance bottom-line. 
% (structure-preserving\slash zero-communication)
}}
}

\subsubsection{Ablation Study on Remote Variants}\label{ssec:results-ablation-arxiv}
\begin{figure}[t]
  \centering
  \includegraphics[width=\linewidth]{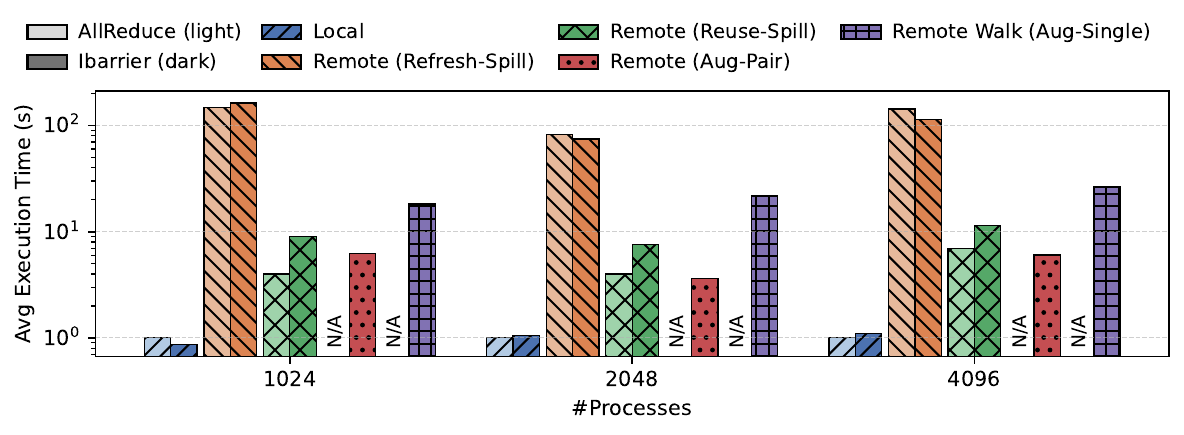}
  \caption{\small End-to-end time of \nomad variants on ogbn-arxiv}
  \label{fig:arxiv}
\end{figure}

Fig.~\ref {fig:arxiv} presents an ablation of the variants of \nomad (using both \texttt{MPI\_Allreduce} and \texttt{MPI\_Ibarrier} based synchronizations, (see Algorithm~\ref{generate-pairs}) for ogbn-arxiv input (the fresh-only variant is excluded since reuse\slash refresh-spill variants are more appropriate).  We observed the same qualitative ordering of variants on other smaller inputs. The results clearly indicate that both refresh-spill and the augmented single walk versions incur the highest execution times: at least an order of magnitude higher than the rest. Refresh-spill generates fresh samples through repeated distributed random walks, requiring global synchronization several times within a batch. The augmented single walk, on the other hand, performs a single, long, distributed traversal (before entering the training loop) to generate the requisite positive sample pairs. All processes collaboratively generate positive training pairs until the slowest process reaches a global target proportional to the total number of batches. Although this design decouples sample generation from training, it introduces a substantial one time synchronization to exchange the remote pairs and materialize the entire set of positive pairs in memory upfront, stalling the training (but mitigates load imbalance within the training loop due to sampling). 
The combination of global synchronization and increased memory pressure makes the augmented single variant slower than reuse-based spilling, though still faster than the remote execution that repeatedly regenerates walks per batch (i.e., refresh-spill). The augmented-pair is an extension of augmented-single; instead of invoking a single long walk, it performs distributed walks for $\sqrt{\mathbb{B}\cdot batch\; size \cdot 2}$ steps where $\mathbb{B}$ is the \#batches. This results in a major reduction in synchronization\slash point-to-point communication for the random sampling portion of Eqn.~\ref{eqn:cost_nomad}, at the expense of extra computation in expanding the set of positive pairs. Consequently, the augmented-pair variant outperforms the rest of the remote variants. Ultimately, the local walk version consistently achieves the least execution time, as it avoids remote communication during sampling, minimizing the load imbalance. Based on these observations, we exclude remote (refresh-spill) and augmented-single variants from the subsequent performance-oriented experiments on larger graphs and configurations.
\begin{figure*}[t]
  \centering
  \includegraphics[width=\linewidth]{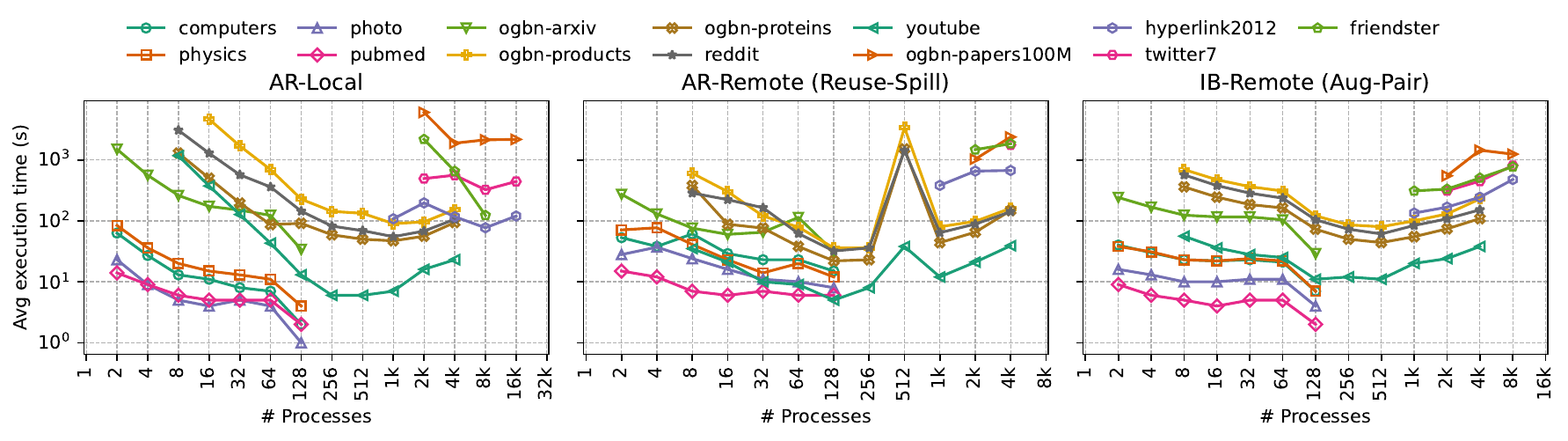}
  \includegraphics[width=\linewidth]{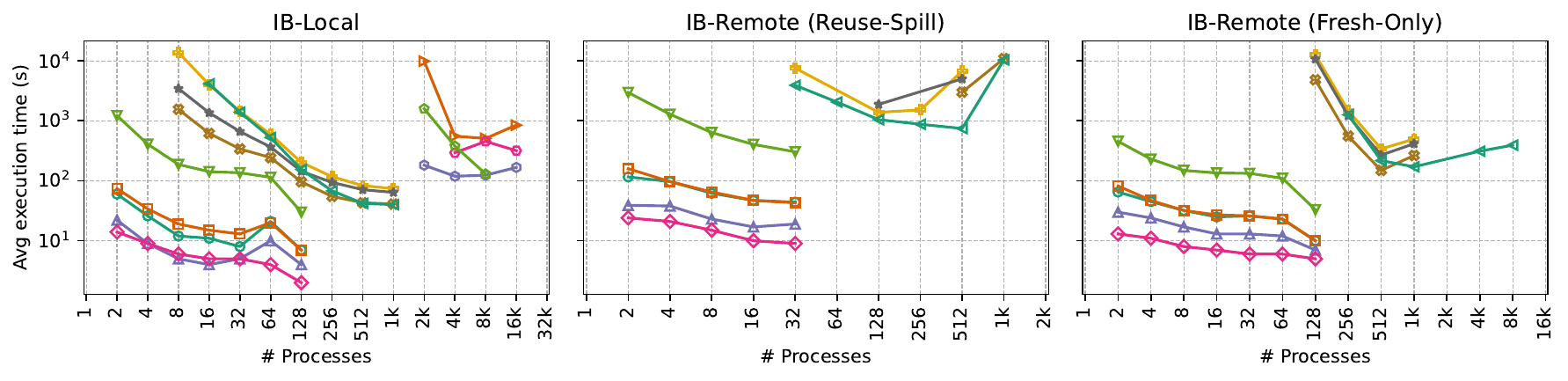}
  \caption{\small Scalability of \nomad on 13 diverse datasets (Table~\ref{tab:datasets}) up to 16K processes. Top: AllReduce; Bottom: Ibarrier.}
  \label{fig:scale}
\end{figure*}
\begin{figure*}[t]
  \centering
  \includegraphics[width=\linewidth]{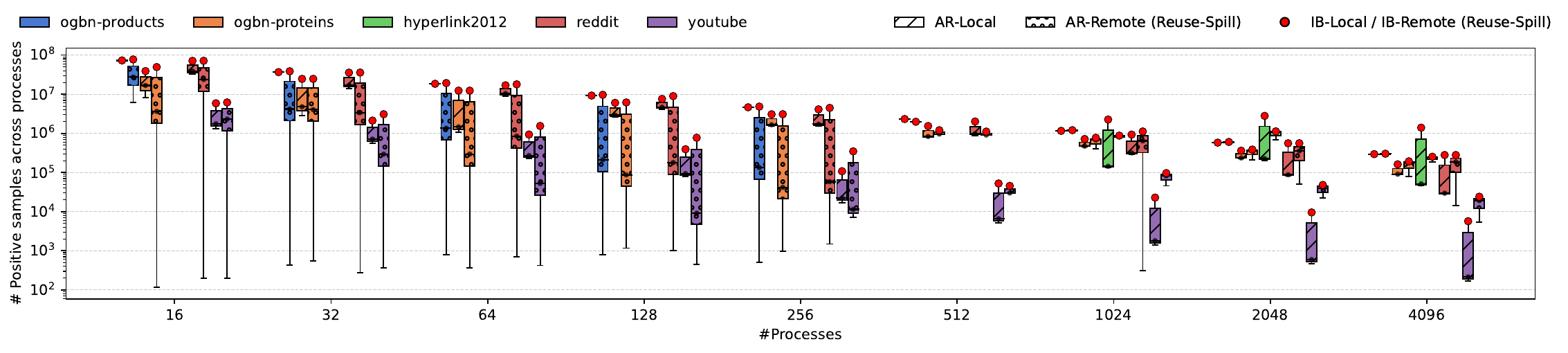}
  \caption{\small Distribution of positive samples across processes for \texttt{AR-Local} and \texttt{AR-Remote (Reuse-Spill)} on large datasets. Red markers denote \texttt{IB-Local} and \texttt{IB-Remote (Reuse-Spill)}, which show little or no variation across processes.}
  \label{fig:load-imbalance}
\end{figure*}
\subsubsection{Scalability vs. Imbalance}\label{ssec:results-scalability}
Fig.~\ref{fig:scale} evaluates the scalability of \nomad (\texttt{AR\slash IB-Local} and remote \texttt{Fresh-Only}, \texttt{Reuse-Spill}, and \texttt{Aug-Pair}) across the datasets in Table~\ref{tab:datasets} and \#processes (using 1--512 nodes; 16--128 processes-per-node). We report average end-to-end execution time, including random sampling and training. Overall, \texttt{AR-Local} delivers the highest speedups: $34$--$44\times$ on computers, physics, pubmed, ogbn-arxiv, and reddit; $25$--$51\times$ on youtube; and $18$--$30\times$ on ogbn-products, ogbn-proteins, and friendster. Even on the largest graphs, where sustained scalability is difficult due to load imbalance, \texttt{AR-Local} remains effective on friendster ($18\times$), but scales poorly on twitter7 ($1.12\times$), ogbn-papers100M ($1.05\times$), and hyperlink2012 ($0.90\times$), indicating worsening imbalance at extreme scale.

% Fig.~\ref{fig:scale} evaluates the scalability of \nomad (\texttt{AR\slash IB-Local} and 3 remote versions \texttt{Fresh-Only}, \texttt{Reuse-Spill}, and \texttt{Aug-Pair}), across graph datasets (see Table~\ref{tab:datasets}) and \#processes (using 1--512 nodes with 16--128 processes-per-node). We report the average end-to-end execution times (including random sampling and training). \texttt{AR-Local} delivers highest speedups in general: $34$--$44\times$ on computers, physics, pubmed, ogbn-arxiv, and reddit, $25$--$51\times$ on youtube, $18$--$30\times$ on ogbn-products, ogbn-proteins, and friendster. Even on the largest graphs where sustainable scalability is challenging due to inherent load imbalance, \texttt{AR-Local} remains effective: it achieves $18\times$ on friendster, although performance remains nearly flat for twitter7 (only $1.12\times$), ogbn-papers ($1.05\times$), and hyperlink ($0.90\times$, slowdown), indicating that load imbalance worsens at extreme scale. 
\begin{table*}[t]
\centering
\caption{\small Communication analysis using CrayPAT on 2048 processes ($n$=32, ppn=64)}
\label{comm-analysis}
\scriptsize
\setlength{\tabcolsep}{4pt}
\renewcommand{\arraystretch}{1.15}

\begin{tabular*}{\textwidth}{@{\extracolsep{\fill}} c c c c c c c c c c c c @{}}
\toprule
\multicolumn{6}{c}{\textbf{arxiv} ($|E|$=2.3M, $\sigma_{E_p}$=643)} &
\multicolumn{6}{c}{\textbf{youtube} ($|E|$=9.8M, $\sigma_{E_p}$=11,127)} \\
\cmidrule(lr){1-6}\cmidrule(lr){7-12}

\multicolumn{3}{c}{\makecell[c]{\textbf{Local} \\ (\%MPI=99.4, Avg.\ Mem./PE=201 MiB)}} &
\multicolumn{3}{c}{\makecell[c]{\textbf{Remote} \\ (\%MPI=99.7, Avg.\ Mem./PE=208 MiB)}} &
\multicolumn{3}{c}{\makecell[c]{\textbf{Local} \\ (\%MPI=99.4, Avg.\ Mem./PE=247 MiB)}} &
\multicolumn{3}{c}{\makecell[c]{\textbf{Remote} \\ (\%MPI=98.8, Avg.\ Mem./PE=260 MiB)}} \\
\cmidrule(lr){1-3}\cmidrule(lr){4-6}\cmidrule(lr){7-9}\cmidrule(lr){10-12}

\makecell[c]{\textbf{\% Total}} &
\makecell[c]{\textbf{\% Imbal.}} &
\textbf{Function} &
\makecell[c]{\textbf{\%Total}} &
\makecell[c]{\textbf{\% Imbal.}} &
\textbf{Function} &
\makecell[c]{\textbf{\% Total}} &
\makecell[c]{\textbf{\% Imbal.}} &
\textbf{Function} &
\makecell[c]{\textbf{\% Total}} &
\makecell[c]{\textbf{\% Imbal.}} &
\textbf{Function} \\
\midrule

46 & 9  & MPI\_Alltoallv &
55 & 12 & MPI\_Allreduce &
76 & 5  & MPI\_Alltoallv &
80 & 1  & MPI\_Alltoallv \\

35 & 45 & MPI File I/O &
15 & 11 & MPI\_Barrier &
11 & 14 & MPI\_Alltoall &
9  & 13 & MPI\_Allreduce \\

12 & 35 & MPI\_Allreduce &
13 & 5  & MPI\_Alltoallv &
9  & 26 & MPI\_Allreduce &
6  & 12 & MPI\_Alltoall \\

7  & 38 & MPI\_Barrier &
9  & 40 & MPI File I/O &
2  & 30 & MPI\_Barrier &
1  & 33 & MPI\_Barrier \\
\bottomrule
\end{tabular*}
\end{table*}
Even \texttt{AR-Local} is not immune to imbalance, as shown in Fig.~\ref{fig:load-imbalance}, which compares the distribution of positive samples across processes for the best local and remote variants. Imbalance is consistently severe for \texttt{AR-Remote (Reuse-Spill)} at low to moderate process counts, with the standard deviation of generated samples reaching three to four orders of magnitude, though it becomes more tolerable beyond 512 processes aside from occasional spikes (e.g., 1024 processes). For \texttt{AR-Local}, the severity depends on the input: ogbn-products shows only 1--2\% variation in generated samples at 4096 processes, whereas youtube remains highly variable, with about $5\times$ and $30\times$ variation at 16 and 4096 processes, respectively (similar trend for hyperlink2012: $30\times$ at 4096 processes). The \textsc{IBarrier} variants (red markers in Fig.~\ref{fig:load-imbalance}) mitigate imbalance by enforcing per-process positive-sample quotas, but increase walk time through stronger straggler effects, as discussed next.
% However, even the ``well-behaving'' variant \texttt{AR-Local} is not immune to load imbalance, as shown in Fig.~\ref{fig:load-imbalance} which compares the distribution of positive samples across processes for the best local\slash remote variants. While the imbalance is consistently severe for \texttt{AR-Remote (Reuse-Spill)} for low-medium \#processes (standard deviation of samples can be three or four orders of magnitude), it becomes tolerable beyond 512 processes, except demonstrating odd spikes (at 1024 process). 
% For\texttt{AR-Local}, the relative degree of imbalance depends on specific inputs. For instance, ogbn-products demonstrates about 1--2\% standard deviation of generated samples across 4096 processes, whereas for youtube standard deviation of samples are highly variable: about 5\slash 30$\times$ on 16\slash 4096 processes (similar trend for hyperlink2012: 30$\times$ on 4096 processes). The \textsc{IBarrier} variants (red markers in Fig.~\ref{fig:load-imbalance}) mitigates load imbalance by enforcing per-process quotas of positive samples, but at the cost of increased walk time due to amplified straggler effects discussed momentarily.

%On the other hand, \textsc{Ibarrier} based synchronization stipulates every process to generate the requisite samples.
\begin{figure}[t]
  \centering
  \includegraphics[width=\linewidth]{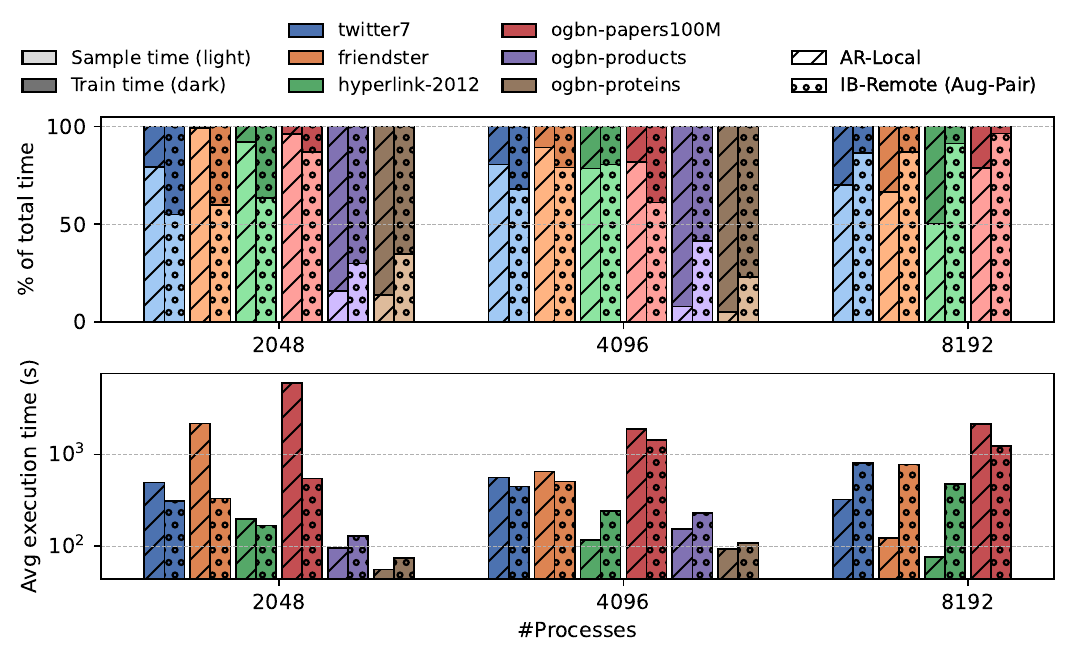}
  \caption{\small Breakdown of end-to-end execution time.}
  \label{fig:breakdown}
\end{figure}

Although Fig.~\ref{fig:load-imbalance} does not fully capture it, several factors can increase the performance variability of \texttt{AR-Remote (Reuse-Spill)}, as seen in Fig.~\ref{fig:scale}. When edge-balanced partitioning fails and execution falls back to the standard 1-D distribution ($|V|/p$, \S\ref{sec:background}), load imbalance can increase, as observed for ogbn-proteins and ogbn-products on 512 processes. Under \textsc{AllReduce}-based synchronization, spill-buffer reuse can then leave some processes with zero positive samples (e.g., ogbn-proteins on 128/256 processes), so they perform no useful work beyond collective communication and barrier waiting. This effect is more damaging to scalability than partition-induced imbalance alone and explains the sharp random-walk slowdowns and runtime variability of Reuse-Spill on ogbn-proteins, ogbn-products, reddit, and youtube in Fig.~\ref{fig:scale}.
% Although it is difficult to be captured by Fig.~\ref{fig:load-imbalance} alone, multitude of possibilities can increase the performance variability for \texttt{AR-Remote (Reuse-Spill)}, as observed in Fig.~\ref{fig:scale}. Failure of edge-balanced graph distribution (falling back to standard 1-D distribution: $|V|/p$, as discussed in \S\ref{sec:overview}) can increase load imbalance, as observed for ogbn-proteins\slash products on 512 processes. Consequently, reusing the spill buffer with \textsc{AllReduce} based synchronization can lead to zero positive samples for an arbitrary process (observed for proteins on 128\slash 256 processes), implying the process does no useful work other than participating in collective communication and waiting on a barrier. This can lead to variabilities in performances at scale (see ogbn-proteins, ogbn-products, reddit, and youtube) for Reuse-Spill due to sharp increase in the random walk times, as seen in Fig.~\ref{fig:scale}. 
\texttt{IB-Aug-Pair} is less sensitive to partition-induced imbalance because it decouples sample generation from training and enforces equal sample counts across processes. This reduces synchronization and communication overhead, yielding better scalability than the other remote variants at the cost of higher memory use. On several datasets, it achieves moderate speedups ($4$--$6\times$ on computers, photo, physics, pubmed, reddit, ogbn-proteins, and ogbn-products, and $8\times$ on ogbn-arxiv), but it does not scale on the largest graphs due to synchronization overheads, including friendster ($0.40\times$), ogbn-papers100M ($0.44\times$), twitter7 ($0.39\times$), hyperlink2012 ($0.28\times$), and youtube ($1.47\times$). However, because \texttt{IB-Aug-Pair} generates synthetic pairs, its performance advantage comes at the cost of lower per-batch sample integrity (positive pairs are no longer tied to a specific exploration). This explains why \texttt{IB-Aug-Pair} achieves the fastest remote execution in Fig.~\ref{fig:scale} while exhibiting subpar embedding quality in Fig.~\ref{fig:quality}.
% Empirically, we observed that such \#samples insufficiency is more inimical to scalability than partition-induced load imbalance alone. \texttt{IB-Aug-Pair} is insensitive to the partitioning-induced imbalance because it decouples sample generation from training and ensures equivalent generated samples. This minimizes the synchronization and communication overhead, resulting in better scalability than other remote variants despite increased memory usage. On several datasets, \texttt{IB-Aug-Pair} achieves moderate speedups (e.g., $4$--$6\times$ on computers, photo, physics, pubmed, reddit, ogbn-proteins, ogbn-products, and $8\times$ on ogbn-arxiv), but does not scale for the largest graphs due to synchronization overheads, including friendster ($0.40\times$), ogbn-papers100M ($0.44\times$), twitter7 ($0.39\times$), hyperlink ($0.28\times$), and youtube ($1.47\times$). However, since it generates synthetic pairs (negative samples or disconnected samples), the performance benefit comes at the cost of subpar per-batch sample integrity, as positive pairs are no longer correlated to a specific exploration. This behavior explains why \texttt{IB-Aug-Pair} achieves the fastest execution times among all remote versions in Fig.~\ref{fig:scale} while simultaneously exhibiting subpar embedding quality in Fig.~\ref{fig:quality}.
In general, Fig.~\ref{fig:scale} shows that \textsc{Ibarrier}-based variants incur higher execution times than their \textsc{AllReduce} counterparts at scale, despite reducing explicit global synchronization and load imbalance (Fig.~\ref{fig:load-imbalance}). For example, \texttt{IB-Remote (Fresh-Only)} shows no scalability on reddit and youtube ($0.38$--$0.60\times$), while \texttt{IB-Remote (Reuse-Spill)} similarly degrades on ogbn-proteins ($0.28\times$). Although processes generate samples asynchronously, random sampling continues until all processes satisfy the predefined non-zero sample quota (\S\ref{method-random-sampling}), increasing communication and imposing additional work even after some processes have already met their quota.
\subsubsection{Timing\slash Communication breakdown}\label{ssec:result-breakdown}
Fig.~\ref{fig:breakdown} decomposes end-to-end time into random sampling and training for six large datasets using the best \nomad variants, \texttt{AR-Local} and \texttt{IB-Remote Aug-Pair}. The balance between sampling and training varies with graph structure (degrees, partitions, etc.) and \nomad variant, but sampling generally dominates execution time. For \texttt{AR-Local}, sampling always generates fresh positive pairs and is dominated by walks over the local partition; as \#processes increases, partitions shrink, reducing sampling cost and bringing it closer to training. In contrast, \texttt{Aug-Pair} pays an upfront distributed communication cost to construct and exchange an initial pool of positive samples, after which per-batch sampling cost is negligible. At lower process counts, this amortization can outperform local sampling, but as \#processes increases, the growing global communication cost eventually offsets that advantage. This crossover is visible in Fig.~\ref{fig:breakdown}: on twitter7 and friendster, local sampling becomes faster than \texttt{Aug-Pair} at 8192 processes, while on hyperlink2012 the crossover occurs between 2048 and 4096 processes. The crossover point depends on graph size and structure: it is not observed for ogbn-papers100M ($|E|=$6.4B) within the evaluated scale, but occurs earlier for medium-sized graphs such as ogbn-products and ogbn-proteins, where \texttt{Aug-Pair} remains slower at higher scales.
We also study the impact of communication considering local and remote (\texttt{Refresh-Spill}, \S\ref{method-remote-variants}) random-walk variants of \nomad with \textsc{AllReduce}-based synchronization using CrayPAT\slash PerfTools on two diverse inputs, arxiv and youtube, at 2048 processes (n=32:ppn=64; arxiv ran for 60 batches, whereas youtube ran for about 5K batches), as shown in Table~\ref{comm-analysis}. On arxiv, local sampling yields comparable sampling and training times, but remote sampling is about $5\times$ more expensive than the rest of training. On youtube, the trend reverses: training costing 8--10$\times$ more than sampling. These results corroborate the cost formulation in Eqn.~\ref{eqn:cost_nomad}.

\subsubsection{Performance Trade-offs}\label{ssec:misc-opt}
\begin{figure}[t]
  \centering
  \includegraphics[width=\linewidth]{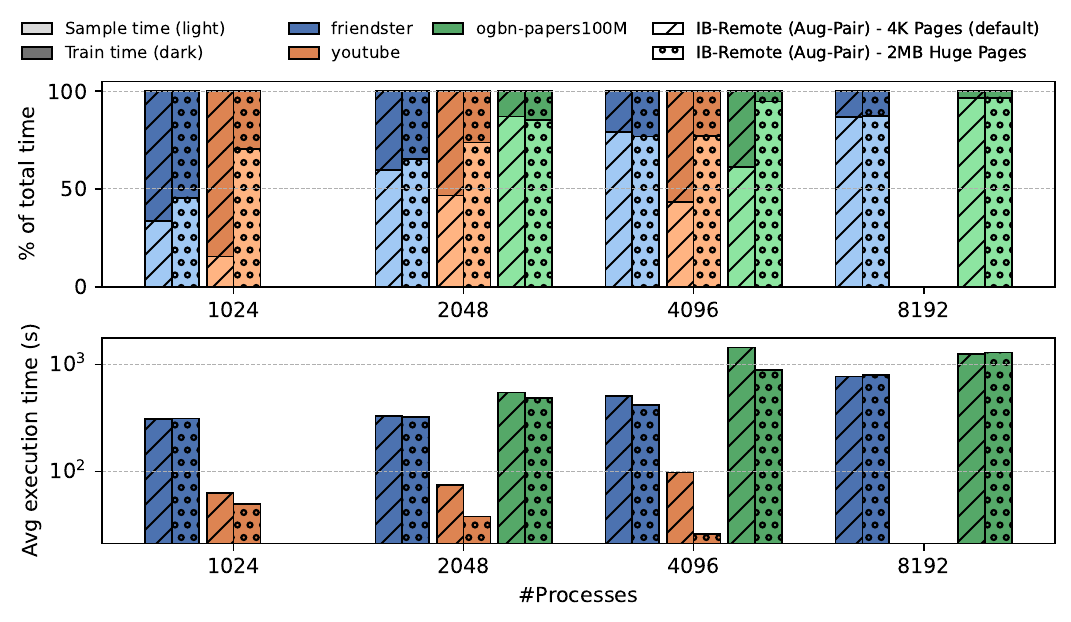}
  \caption{\small 2 MB huge pages on \texttt{IB-Remote (Aug-Pair)}. Top: sampling/training breakdown. Bottom: end-to-end time.}
  \label{fig:hugepages2M}
\end{figure}

On HPE\slash Cray systems, the NIC can track only a limited number of memory pages, so larger pages (2 MB recommended) can improve effective addressability for small- and medium-sized messages. Because huge pages are a node-level resource, availability depends on memory fragmentation, and the runtime may fall back to regular 4 KB pages; this variability is why we exclude huge pages from baseline runs. Fig.~\ref{fig:hugepages2M} evaluates 2 MB huge pages on the best remote variant of \nomad, \texttt{IB-Remote (Aug-Pair)}. Their impact varies by input and scale. On ogbn-papers100M, huge pages improve sampling at 2048 processes for an additional $1.14\times$ speedup, and improve training at 4096 processes for a net $2\times$ reduction in total time, suggesting better memory locality and TLB efficiency during large embedding updates rather than random walks. These benefits diminish at 8192 processes. On youtube, which has smaller per-rank graph partitions but frequent embedding updates (\#epoch=100, batch size=10k), huge pages consistently reduce training time and provide up to $4\times$ speedup on 1024--4096 processes; the default 4 KB-page baseline does not scale over this range (Fig.~\ref{fig:scale}). On friendster, the maximum benefit is 16\% at 4096 processes.
% (size$=\mathcal{O}(|V|/p)\times d$, where $d=$dimension) 

We also develop an MPI-3.0 Remote Memory Access (RMA) variant of \nomad using passive-target synchronization and the unified memory model~\cite{dinan2016implementation}. Unlike the default collective version, which couples communication and synchronization, RMA separates them through one-sided \emph{get} and \emph{accumulate} operations with \texttt{flush} calls for local/remote completion. Local context and vertex embeddings are exposed through MPI windows. After local positive and negative updates (Lines~\ref{alg:defer-pos-remote} and \ref{alg:neg-updates-local}, Algorithm~\ref{alg:nomad_simplified}), a sequence of asynchronous \emph{get} operations fetches the remote context\slash vertex vectors. A local flush completes outstanding gets, enabling the corresponding remote updates (Lines~\ref{alg:line:emb-update-begin}--\ref{alg:line:emb-update-end}) and direct accumulation of remote deltas via RMA accumulate. A subsequent flush ensures remote completion, replacing the \texttt{MPI\_Alltoallv} on Line~\ref{alg:line:mpi-a2a-send-delta}.
\begin{figure}[b]
  \centering 
  \includegraphics[width=\linewidth]{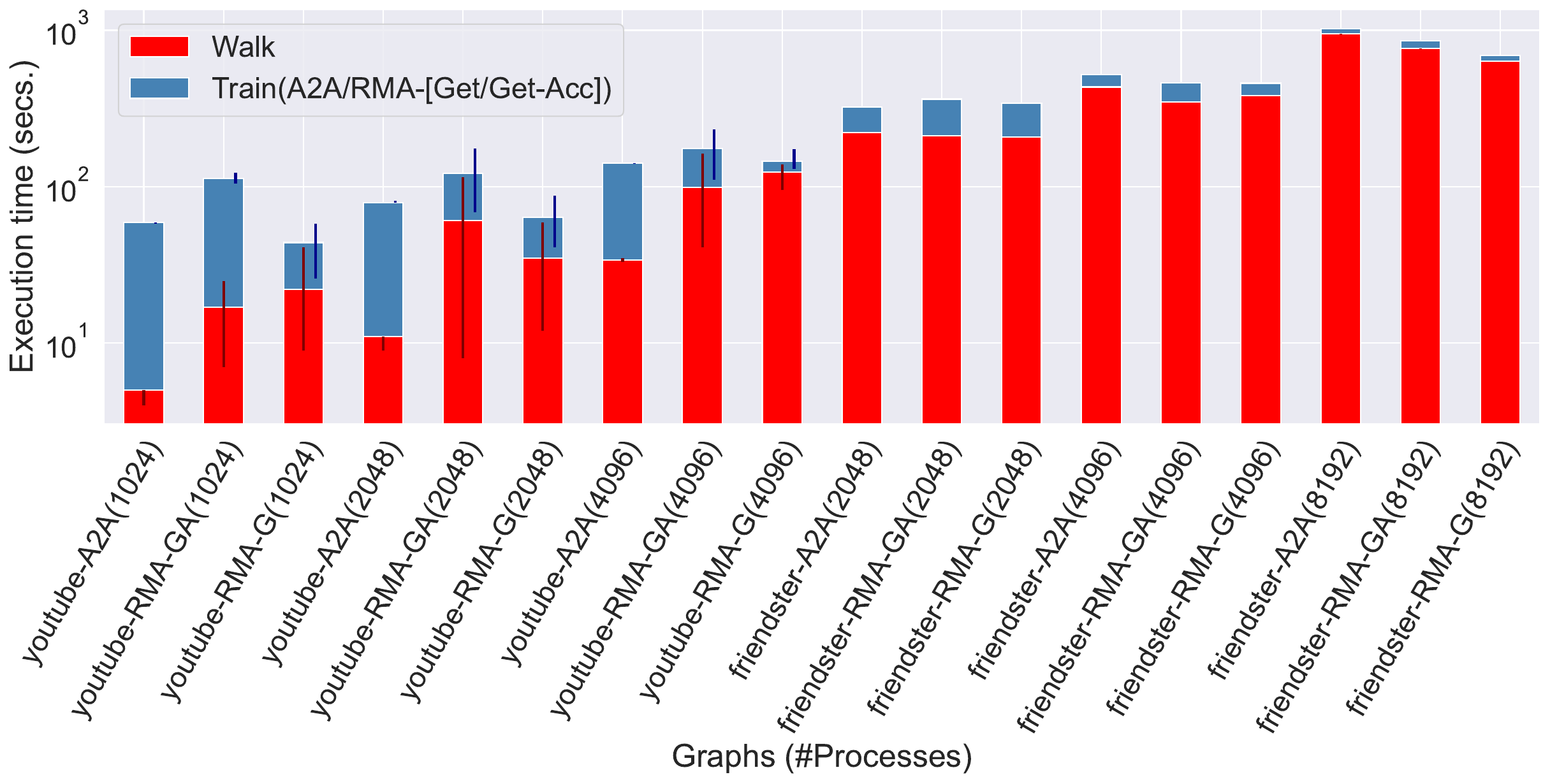}
  \caption{\small RMA variants increases load imbalance for \texttt{Refresh-Spill} (depicted by error bars): comparing regular against RMA for \texttt{Refresh-Spill} and \texttt{Aug-Pair}.}
  \label{fig:rma-overlap-nomad}
\end{figure}

Fig.~\ref{fig:rma-overlap-nomad} shows the performance of RMA variants on youtube (\texttt{AR-Remote Refresh-Spill}) and friendster (\texttt{AR-Remote Aug-Pair}) at thousands of processes (chosen due to irregularities in the edge distributions ($\sigma_{E_p}$ $>$ 5000)). We also evaluate a \emph{get-accumulate} variant (denoted $-GA$) in place of \emph{get} ($-G$) to exploit atomic execution; it improves accuracy by 2--5\% on these inputs, but at additional performance cost. RMA increases load imbalance in constituent routines within the training loop, particularly the repeated sampling phase of \texttt{Refresh-Spill}; in contrast, imbalance is less pronounced for \texttt{Aug-Pair}, where the one-time sampling phase occurs outside the training loop (see \S\ref{method-remote-variants}). Despite this, most RMA configurations improve performance by 10--20\%. 

Table~\ref{rma-overheads} quantifies this imbalance in MPI utilization: while communication dominates in both, the default variant exhibits at most 10\% imbalance, whereas RMA exceeds 50\%. 
% \setlength{\fboxsep}{3pt}
% \noindent\fbox{%
% \parbox{0.97\columnwidth}{%
% \textbf{Takeaway.} Performance depends on how each variant amortizes sample generation under graph irregularity and load imbalance.
% %
% }}
\begin{table}[h]
\centering
\scriptsize
\caption{\small Top \#3 MPI util. ($>$90\% of overall time) for default and RMA variants of \nomad for youtube (1024 processes).}\label{rma-overheads}
\resizebox{\columnwidth}{!}{%
\begin{tabular}{|ccc||ccc|}
\hline
\multicolumn{3}{|c||}{youtube (default) MPI utilization}                                    & \multicolumn{3}{c|}{youtube (RMA) MPI utilization}                                                     \\ \hline \hline
\multicolumn{1}{|c|}{\%-time} & \multicolumn{1}{c|}{\%-imb.} & Function       & \multicolumn{1}{c|}{\%-time} & \multicolumn{1}{c|}{\%-imb.} & Function                    \\ \hline
\multicolumn{1}{|c|}{74}              & \multicolumn{1}{c|}{1}            & MPI\_Alltoallv & \multicolumn{1}{c|}{51}              & \multicolumn{1}{c|}{42}           & MPI\_Allreduce              \\ \hline
\multicolumn{1}{|c|}{15}              & \multicolumn{1}{c|}{10}           & MPI\_Allreduce & \multicolumn{1}{c|}{27}              & \multicolumn{1}{c|}{51}           & MPI\_Win\_flush\_all        \\ \hline
\multicolumn{1}{|c|}{6}               & \multicolumn{1}{c|}{10}           & MPI\_Alltoall  & \multicolumn{1}{c|}{13}              & \multicolumn{1}{c|}{58}           & MPI\_Win\_flush\_local\_all \\ \hline
\end{tabular}%
}
\end{table}
\section{Conclusion}
To the best of our knowledge, \nomad is the first work to study practical performance/quality trade-offs for MPI-based graph embedding on real-world graphs and provides flexible trade-offs among communication, synchronization, and sample quality across shared\slash distributed-memory platforms. Across 13 graphs ranging from millions to billions of edges, \nomad achieves substantial speedups over shared\slash distributed baselines ($370\times$ on small\slash medium graphs, $14$--$22\times$ on billion-edge graphs, and $35$--$76\times$ relative to PBG), with competitive Micro-F1/Macro-F1 relative to LINE, node2vec, and GraphVite. The performance/quality outcomes are driven by the choice of hyperparameters, configurations, and graphs.
\section*{Acknowledgment}
This research is supported by the National Science Foundation (NSF) under Award 2243775 and the U.S. Department of Energy (DOE) through the Office of Advanced Scientific Computing Research's ``Orchestration for Distributed \& Data-Intensive Scientific Exploration", Biological and Environmental Research funded projects ``Orchestrated Platform for Autonomous Laboratories" and ``Foundational AI Models for Optimizing and Understanding Biological Systems" (OPAL FAMOUS) under LAB-25-3560 ``Transformational AI Model Consortium". Pacific Northwest National Laboratory is operated by Battelle for the DOE under Contract DE-AC05-76RL01830. This research used resources of the National Energy Research Scientific Computing Center (NERSC), a Department of Energy User Facility.

\bibliographystyle{IEEEtran}
\bibliography{references}

\end{document}